%% file: main.tex
\documentclass[sigconf,nonacm]{acmart}
\usepackage{popets}

\setcopyright{none}
\usepackage{amsmath}
\usepackage{booktabs}
\usepackage{multirow}
\usepackage{tikz}
\usetikzlibrary{arrows.meta,fit,positioning,backgrounds,calc}

\input{results}

\newcommand{\cmark}{\checkmark}
\newcommand{\xmark}{--}
\newcommand{\mir}{MIRAGE}

\begin{document}

\title{MIRAGE: Motion Identity Reduction via Autoregressive Generative Encoding for Privacy-Preserving Skeleton-based Motion Data}

\author{Thomas Carr$^{1,2}$, Xujun Che$^{1}$, Depeng Xu$^{1,*}$, Shuhan Yuan$^{3}$, Aidong Lu$^{1}$}
\affiliation{%
  \institution{$^{1}$University of North Carolina at Charlotte, Charlotte, NC, USA\\
  $^{2}$Incerta Intelligence, Charlotte, NC, USA\\
  $^{3}$Utah State University, Logan, UT, USA\\[2pt]
  \normalfont\small $\{$tcarr23, xche, dxu7, aidong.lu$\}$@charlotte.edu, thomas.carr@incertaintel.ai, shuhan.yuan@usu.edu}
  \city{}
  \country{}}
\renewcommand{\shortauthors}{Carr et al.}

\begin{abstract}
Skeleton motion identifies people through body proportions and movement style, and skeleton streams are increasingly passed to third-party analytics that run action classifiers trained on raw data. A prior adversarial anonymizer already adds a per-frame residual in the raw coordinate space, but sees one frame at a time; motion retargeting needs a target skeleton and complete sequences, which rules out online use. We present \mir{}, a target-free, causal anonymizer that adds an identity-reducing residual to each incoming frame in the raw coordinate space, computed by attention over past frames so that it can depend on movement over time. A single causal, optionally windowed attention mask governs every attention in a rotary-position encoder--decoder, and key/value caches let the trained function run frame by frame; on all \EqNumSeq{} NTU~60 sequences, batched and streaming outputs differ by at most \EqMaxAbs\,m. Our reported results use an unbounded window, for which per-stream state grows with stream length; bounded state requires a finite window, which changes the privacy--utility operating point. We state a threat model with a passive adversary that applies raw-trained identity models and an adaptive adversary that trains on the output of the released anonymizer, treat the adaptive adversary as the headline, and evaluate on NTU RGB+D~60 over five generator seeds, alongside the per-frame residual anonymizer of Moon et al.\ trained under the same protocol.
When consumers retrain on the released data, \mir{} keeps \RetSgnAR\% action recognition (raw \RawSgnAR\%; Moon et al.\ \MoonRetSgnAR\%) while a retrained identity classifier reaches \RetSgnRI\% re-identification (RI) over \NumSubjects{} subjects, against \MoonRetSgnRI\% for Moon et al.\ and \RawSgnRI\% on raw data. The strongest adaptive attacker, \WorstRetAttacker{}, still reaches \WorstRetRI\% (chance \ChanceRI\%), so \mir{} reduces identity leakage; it does not anonymize. Against raw-trained (frozen) classifiers, \mir{} lowers RI below the strongest retargeting baseline (\MainSgnRI\% vs.\ \TmrSgnRI\%) but frozen action recognition falls to \MainSgnAR\% (vs.\ \TmrSgnAR\%; all Welch $p<0.05$): the output is shifted away from the distribution raw-trained classifiers expect, so at full residual scale \mir{} is not a drop-in replacement for raw input to existing frozen models.
\end{abstract}

\keywords{skeleton motion, de-identification, biometric privacy, streaming, re-identification, adaptive adversary}

\maketitle
{\renewcommand{\thefootnote}{$*$}\footnotetext{Corresponding author: Depeng Xu (dxu7@charlotte.edu).}}

\input{sec_intro}

\input{sec_related}
\input{sec_threat}
\input{sec_method}
\input{sec_eval}
\input{sec_results}
\input{sec_discussion}

\appendix

\begin{ethics}
This work performs secondary analysis of previously released research datasets (NTU RGB+D~60 and~120 and ETRI-Activity3D) under their existing access terms. We did not recruit participants, collect new data, interact with dataset participants, or attempt to link the pseudonymous subject identifiers in these datasets to real people. The experiments train and run identity classifiers in order to measure residual identification risk; we report only aggregate and pseudonymous per-subject statistics and will not release identity-classifier outputs tied to individual recordings or any raw biometric samples. The work was not submitted to an ethics board because it uses existing research artifacts without participant interaction. Skeleton de-identification is dual use: it can protect people whose motion is shared with analytics services, and it could also be used to frustrate lawful identification such as forensic gait analysis. Our results show that identity remains recoverable by an adaptive adversary, so the method must not be presented to data subjects as anonymization in a legal or technical sense; deployments on non-public data need their own review of consent, purpose limitation, retention, and access control.
\end{ethics}

\begin{openscience}
We will release the \mir{} source code, training and generation configurations for every reported run, the evaluation scripts (frozen and retrained evaluators, gait attackers, 1-NN and linkage attacks, per-subject analysis, significance tests, streaming benchmark), the unit tests that pin prefix invariance, batch--stream equality, bounded caches past the training length, and padding independence, and the trained generator checkpoints. The attacker adapters we wrote for third-party gait models will be released with their upstream licenses. The datasets and the third-party classifier code remain available only from their owners under their licenses; the release documents the expected versions.
\end{openscience}

\begin{ai}
The authors used generative AI-based tools to revise the manuscript, improve flow and grammar, check that the methods section matches the implementation, audit the bibliography against publisher records, and assist with evaluation, analysis and plotting scripts. Figures are deterministic renderings of recorded experiment artifacts; AI-based tools did not supply generative imagery or experimental measurements. We have manually verified and are responsible for the accuracy, originality, and integrity of the output of all AI-based tools.
\end{ai}

\bibliographystyle{ACM-Reference-Format}
\bibliography{combined}

\input{appendix}

\end{document}

%% file: results.tex
\newif\ifdraftresults
\draftresultsfalse
\newif\iffinalresults
\finalresultstrue

\newcounter{resultsvone}
\newcounter{resultstbd}

\DeclareRobustCommand{\vtwo}[1]{#1}
\DeclareRobustCommand{\keep}[1]{#1}

\AtEndDocument{%
  \PackageWarningNoLine{results.tex}{%
    \arabic{resultsvone}\space stale v1 value(s) and
    \arabic{resultstbd}\space TBD value(s) are typeset}}

\newcommand{\setres}[2]{\expandafter\def\csname res:#1\endcsname{#2}}
\newcommand{\res}[1]{%
  \ifcsname res:#1\endcsname\csname res:#1\endcsname
  \else\errmessage{results.tex: undefined result key '#1'}\fi}

\newcommand{\pmsd}[2]{#1{\scriptsize$\,\pm\,$#2}}

\newcommand{\ChanceRI}{2.5}          %
\newcommand{\ChanceAR}{2.0}          %
\newcommand{\ChanceLink}{50}         %
\newcommand{\NumSubjects}{40}
\newcommand{\NumGenSeeds}{\vtwo{5}}  %
\newcommand{\SuiteSeeds}{\vtwo{42}}  %
\newcommand{\BenchGPU}{\vtwo{V100S}}  %

\newcommand{\RawSgnAR}{\keep{89.1}}
\newcommand{\RawSgnRI}{\keep{75.4}}
\newcommand{\RawMixAR}{\keep{88.6}}
\newcommand{\RawMixRI}{\keep{72.8}}
\newcommand{\NoiseSgnAR}{\keep{7.3}}
\newcommand{\NoiseSgnRI}{\keep{21.8}}
\newcommand{\NoiseMixAR}{\keep{29.4}}
\newcommand{\NoiseMixRI}{\keep{14.0}}
\newcommand{\NoiseMPJPE}{\keep{0.108}}
\newcommand{\NoiseBone}{\keep{0.062}}
\newcommand{\NoiseSmooth}{\keep{0.271}}
\newcommand{\NoiseVel}{\keep{0.010}}
\newcommand{\DmrSgnAR}{\keep{5.0}}
\newcommand{\DmrSgnRI}{\keep{14.2}}
\newcommand{\DmrMixAR}{\keep{5.2}}
\newcommand{\DmrMixRI}{\keep{14.0}}
\newcommand{\DmrMPJPE}{\keep{0.202}}
\newcommand{\DmrBone}{\keep{0.029}}
\newcommand{\DmrSmooth}{\keep{0.023}}
\newcommand{\DmrVel}{\keep{0.001}}
\newcommand{\PmrSgnAR}{\keep{2.4}}
\newcommand{\PmrSgnRI}{\keep{7.3}}
\newcommand{\PmrMixAR}{\keep{2.2}}
\newcommand{\PmrMixRI}{\keep{9.2}}
\newcommand{\PmrMPJPE}{\keep{0.272}}
\newcommand{\PmrBone}{\keep{0.051}}
\newcommand{\PmrSmooth}{\keep{0.094}}
\newcommand{\PmrVel}{\keep{0.002}}
\newcommand{\TmrSgnAR}{\keep{75.82}}
\newcommand{\TmrSgnARsd}{\keep{4.64}}
\newcommand{\TmrSgnRI}{\keep{22.09}}
\newcommand{\TmrSgnRIsd}{\keep{3.24}}
\newcommand{\TmrMixAR}{\keep{66.50}}
\newcommand{\TmrMixARsd}{\keep{7.40}}
\newcommand{\TmrMixRI}{\keep{17.08}}
\newcommand{\TmrMixRIsd}{\keep{3.97}}
\newcommand{\TmrBestSgnAR}{\keep{77.05}}
\newcommand{\TmrBestSgnRI}{\keep{16.73}}
\newcommand{\TmrBestMixAR}{\keep{71.57}}
\newcommand{\TmrBestMixRI}{\keep{15.21}}
\newcommand{\TmrMPJPE}{\keep{0.211}}
\newcommand{\TmrBone}{\keep{0.007}}
\newcommand{\TmrSmooth}{\keep{0.020}}
\newcommand{\TmrVel}{\keep{0.001}}
\newcommand{\MainSgnAR}{\vtwo{37.04}}  %
\newcommand{\MainSgnARsd}{\vtwo{4.85}}  %
\newcommand{\MainSgnRI}{\vtwo{10.36}}  %
\newcommand{\MainSgnRIsd}{\vtwo{1.10}}  %
\newcommand{\MainMixAR}{\vtwo{39.98}}  %
\newcommand{\MainMixARsd}{\vtwo{3.66}}  %
\newcommand{\MainMixRI}{\vtwo{9.44}}  %
\newcommand{\MainMixRIsd}{\vtwo{1.04}}  %
\newcommand{\MainSgnARmin}{\vtwo{29.7}}  %
\newcommand{\MainSgnARmax}{\vtwo{42.0}}  %
\newcommand{\MainMPJPE}{\vtwo{0.209}}  %
\newcommand{\MainBone}{\vtwo{0.039}}  %
\newcommand{\MainSmooth}{\vtwo{0.053}}  %
\newcommand{\MainVel}{\vtwo{0.001}}  %
\newcommand{\AoneSgnAR}{\vtwo{34.3}}  %
\newcommand{\AoneSgnRI}{\vtwo{9.9}}  %
\newcommand{\AoneMixAR}{\vtwo{34.4}}  %
\newcommand{\AoneMixRI}{\vtwo{8.8}}  %
\newcommand{\AoneMPJPE}{\vtwo{0.204}}  %
\newcommand{\AoneBone}{\vtwo{0.035}}  %
\newcommand{\AoneSmooth}{\vtwo{0.052}}  %
\newcommand{\AoneVel}{\vtwo{0.001}}  %
\newcommand{\BetaSgnAR}{\vtwo{39.8}}  %
\newcommand{\BetaSgnRI}{\vtwo{11.3}}  %
\newcommand{\BetaMixAR}{\vtwo{40.3}}  %
\newcommand{\BetaMixRI}{\vtwo{10.0}}  %
\newcommand{\BetaMPJPE}{\vtwo{0.184}}  %
\newcommand{\BetaBone}{\vtwo{0.032}}  %
\newcommand{\BetaSmooth}{\vtwo{0.049}}  %
\newcommand{\BetaVel}{\vtwo{0.001}}  %

\newcommand{\MoonNumSeeds}{\vtwo{3}}  %
\newcommand{\MoonSgnAR}{\vtwo{88.70}}  %
\newcommand{\MoonSgnARsd}{\vtwo{0.17}}  %
\newcommand{\MoonSgnRI}{\vtwo{44.27}}  %
\newcommand{\MoonSgnRIsd}{\vtwo{5.42}}  %
\newcommand{\MoonMixAR}{\vtwo{86.23}}  %
\newcommand{\MoonMixARsd}{\vtwo{0.57}}  %
\newcommand{\MoonMixRI}{\vtwo{33.20}}  %
\newcommand{\MoonMixRIsd}{\vtwo{4.78}}  %
\newcommand{\MoonMPJPE}{\vtwo{0.087}}  %
\newcommand{\MoonBone}{\vtwo{0.027}}  %
\newcommand{\MoonSmooth}{\vtwo{0.036}}  %
\newcommand{\MoonVel}{\vtwo{$<$0.001}}  %

\newcommand{\SigSgnARdiff}{\vtwo{\ensuremath{-}38.78}}  %
\newcommand{\SigSgnARlo}{\vtwo{\ensuremath{-}45.70}}  %
\newcommand{\SigSgnARhi}{\vtwo{\ensuremath{-}31.86}}  %
\newcommand{\SigSgnARp}{\vtwo{\ensuremath{1.2\times10^{-6}}}}  %
\newcommand{\SigSgnARg}{\vtwo{\ensuremath{-}7.39}}  %
\newcommand{\SigSgnRIdiff}{\vtwo{\ensuremath{-}11.73}}  %
\newcommand{\SigSgnRIlo}{\vtwo{\ensuremath{-}15.68}}  %
\newcommand{\SigSgnRIhi}{\vtwo{\ensuremath{-}7.77}}  %
\newcommand{\SigSgnRIp}{\vtwo{0.00065}}  %
\newcommand{\SigSgnRIg}{\vtwo{\ensuremath{-}4.38}}  %
\newcommand{\SigMixARdiff}{\vtwo{\ensuremath{-}26.52}}  %
\newcommand{\SigMixARlo}{\vtwo{\ensuremath{-}35.61}}  %
\newcommand{\SigMixARhi}{\vtwo{\ensuremath{-}17.43}}  %
\newcommand{\SigMixARp}{\vtwo{0.00041}}  %
\newcommand{\SigMixARg}{\vtwo{\ensuremath{-}4.10}}  %
\newcommand{\SigMixRIdiff}{\vtwo{\ensuremath{-}7.64}}  %
\newcommand{\SigMixRIlo}{\vtwo{\ensuremath{-}12.50}}  %
\newcommand{\SigMixRIhi}{\vtwo{\ensuremath{-}2.78}}  %
\newcommand{\SigMixRIp}{\vtwo{0.011}}  %
\newcommand{\SigMixRIg}{\vtwo{\ensuremath{-}2.38}}  %

\newcommand{\DmrRetSgnAR}{\keep{43.1}}
\newcommand{\DmrRetSgnRI}{\keep{38.1}}
\newcommand{\DmrRetMixAR}{\keep{45.3}}
\newcommand{\DmrRetMixRI}{\keep{38.7}}
\newcommand{\PmrRetSgnAR}{\keep{19.9}}
\newcommand{\PmrRetSgnRI}{\keep{24.2}}
\newcommand{\PmrRetMixAR}{\keep{20.8}}
\newcommand{\PmrRetMixRI}{\keep{25.5}}
\newcommand{\TmrRetSgnAR}{\keep{55.4}}
\newcommand{\TmrRetSgnRI}{\keep{12.2}}
\newcommand{\TmrRetMixAR}{\keep{55.7}}
\newcommand{\TmrRetMixRI}{\keep{12.5}}
\newcommand{\MoonRetSgnAR}{\vtwo{88.7}}  %
\newcommand{\MoonRetSgnARsd}{\vtwo{0.1}}  %
\newcommand{\MoonRetSgnRI}{\vtwo{69.4}}  %
\newcommand{\MoonRetSgnRIsd}{\vtwo{1.8}}  %
\newcommand{\MoonRetMixAR}{\vtwo{88.6}}  %
\newcommand{\MoonRetMixRI}{\vtwo{67.6}}  %
\newcommand{\RetSgnAR}{\vtwo{83.9}}  %
\newcommand{\RetSgnARsd}{\vtwo{0.3}}  %
\newcommand{\RetSgnRI}{\vtwo{43.4}}  %
\newcommand{\RetSgnRIsd}{\vtwo{1.2}}  %
\newcommand{\RetEvalSd}{\vtwo{0.4}}  %
\newcommand{\RetMixAR}{\vtwo{83.5}}  %
\newcommand{\RetMixRI}{\vtwo{48.1}}  %
\newcommand{\AdaptSgnRI}{\vtwo{42.0}}  %
\newcommand{\AdaptMixRI}{\vtwo{49.3}}  %
\newcommand{\GaitGraphRawRI}{\keep{75.9}}
\newcommand{\GaitMixerRawRI}{\keep{54.6}}
\newcommand{\GaitGraphRetRI}{\vtwo{42.4}}  %
\newcommand{\GaitMixerRetRI}{\vtwo{25.7}}  %
\newcommand{\WorstRetRI}{\vtwo{49.3}}  %
\newcommand{\WorstRetAttacker}{\vtwo{adaptive MixFormer}}  %
\newcommand{\WorstRetRawRI}{\keep{72.8}}  %
\newcommand{\NumAtwoAttackers}{six}

\newcommand{\SuiteSgnRawAR}{\keep{89.1}}
\newcommand{\SuiteSgnRawRI}{\keep{75.4}}
\newcommand{\SuiteSgnAR}{\vtwo{29.7}}  %
\newcommand{\SuiteSgnRI}{\vtwo{8.7}}  %
\newcommand{\TailWorstSgn}{\vtwo{41.2}}  %
\newcommand{\TailMedSgn}{\vtwo{0.7}}  %
\newcommand{\SuiteMixRawAR}{\keep{88.6}}
\newcommand{\SuiteMixRawRI}{\keep{72.8}}
\newcommand{\SuiteMixAR}{\vtwo{34.1}}  %
\newcommand{\SuiteMixRI}{\vtwo{8.4}}  %
\newcommand{\TailWorstMix}{\vtwo{24.4}}  %
\newcommand{\TailMedMix}{\vtwo{0.5}}  %
\newcommand{\SuiteCtrRawAR}{\vtwo{94.1}}  %
\newcommand{\SuiteCtrRawRI}{\vtwo{77.2}}  %
\newcommand{\SuiteCtrAR}{\vtwo{39.7}}  %
\newcommand{\SuiteCtrRI}{\vtwo{10.0}}  %
\newcommand{\TailWorstCtr}{\vtwo{35.2}}  %
\newcommand{\TailMedCtr}{\vtwo{2.2}}  %
\newcommand{\SuiteStgRawAR}{\vtwo{92.7}}  %
\newcommand{\SuiteStgRawRI}{\vtwo{72.0}}  %
\newcommand{\SuiteStgAR}{\vtwo{42.3}}  %
\newcommand{\SuiteStgRI}{\vtwo{9.5}}  %
\newcommand{\TailWorstStg}{\vtwo{43.9}}  %
\newcommand{\TailMedStg}{\vtwo{1.6}}  %
\newcommand{\SuiteAgcnRawAR}{\vtwo{93.3}}  %
\newcommand{\SuiteAgcnRawRI}{\vtwo{72.1}}  %
\newcommand{\SuiteAgcnAR}{\vtwo{45.4}}  %
\newcommand{\SuiteAgcnRI}{\vtwo{9.8}}  %
\newcommand{\TailWorstAgcn}{\vtwo{40.5}}  %
\newcommand{\TailMedAgcn}{\vtwo{1.3}}  %
\newcommand{\SuiteMsgRawAR}{\vtwo{83.5}}  %
\newcommand{\SuiteMsgRawRI}{\vtwo{54.3}}  %
\newcommand{\SuiteMsgAR}{\vtwo{45.0}}  %
\newcommand{\SuiteMsgRI}{\vtwo{9.5}}  %
\newcommand{\TailWorstMsg}{\vtwo{29.7}}  %
\newcommand{\TailMedMsg}{\vtwo{0.5}}  %
\newcommand{\SuiteHdRawAR}{\vtwo{81.0}}  %
\newcommand{\SuiteHdRawRI}{\vtwo{48.2}}  %
\newcommand{\SuiteHdAR}{\vtwo{38.6}}  %
\newcommand{\SuiteHdRI}{\vtwo{9.0}}  %
\newcommand{\TailWorstHd}{\vtwo{34.3}}  %
\newcommand{\TailMedHd}{\vtwo{0.8}}  %
\newcommand{\SuiteInfoRawAR}{\vtwo{91.1}}  %
\newcommand{\SuiteInfoRawRI}{\vtwo{67.5}}  %
\newcommand{\SuiteInfoAR}{\vtwo{47.3}}  %
\newcommand{\SuiteInfoRI}{\vtwo{8.1}}  %
\newcommand{\TailWorstInfo}{\vtwo{34.9}}  %
\newcommand{\TailMedInfo}{\vtwo{1.8}}  %
\newcommand{\SuiteFreqRawAR}{\vtwo{83.4}}  %
\newcommand{\SuiteFreqRawRI}{\vtwo{57.6}}  %
\newcommand{\SuiteFreqAR}{\vtwo{41.3}}  %
\newcommand{\SuiteFreqRI}{\vtwo{10.1}}  %
\newcommand{\TailWorstFreq}{\vtwo{35.2}}  %
\newcommand{\TailMedFreq}{\vtwo{0.0}}  %
\newcommand{\SuiteSkateRawAR}{\vtwo{85.3}}  %
\newcommand{\SuiteSkateRawRI}{\vtwo{61.9}}  %
\newcommand{\SuiteSkateAR}{\vtwo{43.9}}  %
\newcommand{\SuiteSkateRI}{\vtwo{9.0}}  %
\newcommand{\TailWorstSkate}{\vtwo{34.8}}  %
\newcommand{\TailMedSkate}{\vtwo{1.0}}  %
\newcommand{\SuiteGaitGraphRI}{\vtwo{10.1}}  %
\newcommand{\TailWorstGaitGraph}{\vtwo{34.3}}  %
\newcommand{\TailMedGaitGraph}{\vtwo{1.8}}  %
\newcommand{\SuiteGaitMixerRI}{\vtwo{6.5}}  %
\newcommand{\TailWorstGaitMixer}{\vtwo{16.0}}  %
\newcommand{\TailMedGaitMixer}{\vtwo{1.0}}  %
\newcommand{\SuiteMeanAR}{\vtwo{40.7}}  %
\newcommand{\SuiteMeanRI}{\vtwo{9.2}}  %
\newcommand{\SuiteMinRI}{\vtwo{8.1}}  %
\newcommand{\SuiteMaxRI}{\vtwo{10.1}}  %
\newcommand{\KnnTopOne}{\vtwo{9.1}}  %
\newcommand{\KnnRecordLink}{\vtwo{9.2}}  %
\newcommand{\LinkWithinAcc}{\vtwo{50.2}}  %
\newcommand{\LinkWithinAUC}{\vtwo{0.506}}  %
\newcommand{\LinkCrossAcc}{\vtwo{50.2}}  %
\newcommand{\LinkCrossAUC}{\vtwo{0.502}}  %
\newcommand{\LinkRawAcc}{\keep{67}}
\newcommand{\TailRatioMin}{\vtwo{2.4}}     %
\newcommand{\TailRatioMax}{\vtwo{4.7}}     %
\newcommand{\TailWorstMax}{\vtwo{43.9}}     %
\newcommand{\TailWorstMaxEval}{\vtwo{ST-GCN}} %
\newcommand{\TailMedMin}{\vtwo{0.0}}       %
\newcommand{\TailMedMax}{\vtwo{2.2}}  %
\newcommand{\TailSameTop}{\vtwo{7}}      %

\newcommand{\EqNumSeq}{\vtwo{56{,}578}}  %
\newcommand{\EqMaxAbs}{\vtwo{\ensuremath{7.2\times10^{-6}}}}  %
\newcommand{\EqMeanAbs}{\vtwo{\ensuremath{7.8\times10^{-8}}}}  %
\newcommand{\EqBatchSgnAR}{\vtwo{29.7}}  %
\newcommand{\EqBatchSgnRI}{\vtwo{8.7}}  %
\newcommand{\EqStreamSgnAR}{\vtwo{29.7}}  %
\newcommand{\EqStreamSgnRI}{\vtwo{8.7}}  %
\newcommand{\StreamMsFrame}{\vtwo{8.22}}  %
\newcommand{\StreamFPS}{\vtwo{122}}  %
\newcommand{\StreamPninetyfive}{\vtwo{8.36}}  %
\newcommand{\StreamPeakMB}{\vtwo{42.6}}  %
\newcommand{\BatchMsFrame}{\vtwo{0.14}}  %
\newcommand{\BatchFPS}{\vtwo{7{,}116}}  %
\newcommand{\BatchPeakMB}{\vtwo{41.8}}  %
\newcommand{\LongT}{\vtwo{100{,}000}}  %
\newcommand{\LongW}{\vtwo{64}}  %
\newcommand{\LongCacheEntries}{\vtwo{64}}  %
\newcommand{\LongCacheKB}{\vtwo{1{,}536}}  %
\newcommand{\LongMsStart}{\vtwo{7.61}}  %
\newcommand{\LongMsEnd}{\vtwo{7.55}}  %
\newcommand{\LongPeakMBStart}{\vtwo{42.6}}  %
\newcommand{\LongPeakMBEnd}{\vtwo{42.6}}  %
\newcommand{\LongInfT}{\vtwo{10{,}000}}  %
\newcommand{\LongBenchGPU}{\vtwo{A40}}  %
\newcommand{\CostBenchGPU}{\vtwo{A40}}  %
\newcommand{\LongInfPeakMBEnd}{\vtwo{334.3}}  %
\newcommand{\LongInfMsEnd}{\vtwo{16.2}}  %

\newcommand{\CostMirageParams}{\vtwo{7.43}}
\newcommand{\CostTmrParams}{\keep{23.7}}
\newcommand{\CostDmrParams}{\keep{4.9}}
\newcommand{\CostPmrParams}{\keep{1.0}}
\newcommand{\CostParamRatio}{\vtwo{3.2}}  %
\newcommand{\CostMirageGflopsSeq}{\vtwo{1.00}}  %
\newcommand{\CostTmrMsSeq}{\keep{458}}       %
\newcommand{\CostDmrMsSeq}{\keep{1.1}}
\newcommand{\CostPmrMsSeq}{\keep{1.2}}
\newcommand{\CostMirageStreamMsSeq}{\vtwo{465}}  %
\newcommand{\CostMirageNoCacheMsSeq}{\vtwo{502}}  %
\newcommand{\CostMirageBatchMsSeq}{\vtwo{8.08}}  %
\newcommand{\CostMirageTrainH}{\vtwo{3.85}} %
\newcommand{\CostTmrTrainH}{\keep{28}}
\newcommand{\CostDmrTrainH}{\keep{3}}
\newcommand{\CostPmrTrainH}{\keep{4}}

\newcommand{\AblFullFrzAR}{\vtwo{29.7}}  %
\newcommand{\AblFullFrzRI}{\vtwo{8.7}}  %
\newcommand{\AblFullRetAR}{\vtwo{83.7}}  %
\newcommand{\AblFullRetRI}{\vtwo{41.9}}  %
\newcommand{\AblNoConFrzAR}{\vtwo{39.0}}  %
\newcommand{\AblNoConFrzRI}{\vtwo{10.7}}  %
\newcommand{\AblNoConRetAR}{\vtwo{84.2}}  %
\newcommand{\AblNoConRetRI}{\vtwo{45.1}}  %
\newcommand{\AblNoDdFrzAR}{\vtwo{24.3}}  %
\newcommand{\AblNoDdFrzRI}{\vtwo{9.9}}  %
\newcommand{\AblNoDdRetAR}{\vtwo{84.7}}  %
\newcommand{\AblNoDdRetRI}{\vtwo{47.3}}  %
\newcommand{\AblNeitherFrzAR}{\vtwo{23.4}}  %
\newcommand{\AblNeitherFrzRI}{\vtwo{9.8}}  %
\newcommand{\AblNeitherRetAR}{\vtwo{84.4}}  %
\newcommand{\AblNeitherRetRI}{\vtwo{48.0}}  %
\newcommand{\AblNoOutIdFrzAR}{\vtwo{56.0}}  %
\newcommand{\AblNoOutIdFrzRI}{\vtwo{16.3}}  %
\newcommand{\AblNoOutIdRetAR}{\vtwo{84.9}}  %
\newcommand{\AblNoOutIdRetRI}{\vtwo{57.9}}  %
\newcommand{\AblNoActFrzAR}{\vtwo{74.2}}  %
\newcommand{\AblNoActFrzRI}{\vtwo{19.7}}  %
\newcommand{\AblNoActRetAR}{\vtwo{84.6}}  %
\newcommand{\AblNoActRetRI}{\vtwo{57.2}}  %
\newcommand{\AblNoLatIdFrzAR}{\vtwo{41.4}}  %
\newcommand{\AblNoLatIdFrzRI}{\vtwo{9.8}}  %
\newcommand{\AblNoLatIdRetAR}{\vtwo{84.7}}  %
\newcommand{\AblNoLatIdRetRI}{\vtwo{45.3}}  %
\newcommand{\AblNoTempFrzAR}{\vtwo{37.9}}  %
\newcommand{\AblNoTempFrzRI}{\vtwo{10.6}}  %
\newcommand{\AblNoTempRetAR}{\vtwo{83.9}}  %
\newcommand{\AblNoTempRetRI}{\vtwo{44.5}}  %
\newcommand{\AblNoBoneFrzAR}{\vtwo{45.4}}  %
\newcommand{\AblNoBoneFrzRI}{\vtwo{10.4}}  %
\newcommand{\AblNoBoneRetAR}{\vtwo{83.9}}  %
\newcommand{\AblNoBoneRetRI}{\vtwo{43.4}}  %
\newcommand{\AblNoProgFrzAR}{\vtwo{41.4}}  %
\newcommand{\AblNoProgFrzRI}{\vtwo{9.1}}  %
\newcommand{\AblNoProgRetAR}{\vtwo{83.8}}  %
\newcommand{\AblNoProgRetRI}{\vtwo{43.6}}  %

\newcommand{\AlphaPtOneRetAR}{\vtwo{84.8}}  %
\newcommand{\AlphaPtOneRetRI}{\vtwo{47.2}}  %
\newcommand{\AlphaPtOneFrzAR}{\vtwo{41.1}}  %
\newcommand{\AlphaPtOneFrzRI}{\vtwo{11.0}}  %
\newcommand{\AlphaPtTwoFiveRetAR}{\vtwo{84.3}}  %
\newcommand{\AlphaPtTwoFiveRetRI}{\vtwo{45.5}}  %
\newcommand{\AlphaPtTwoFiveFrzAR}{\vtwo{41.0}}  %
\newcommand{\AlphaPtTwoFiveFrzRI}{\vtwo{11.4}}  %
\newcommand{\AlphaPtFiveRetAR}{\vtwo{83.7}}  %
\newcommand{\AlphaPtFiveRetRI}{\vtwo{41.9}}  %
\newcommand{\AlphaPtFiveFrzAR}{\vtwo{29.7}}  %
\newcommand{\AlphaPtFiveFrzRI}{\vtwo{8.7}}  %
\newcommand{\AlphaPtSevenFiveRetAR}{\vtwo{83.5}}  %
\newcommand{\AlphaPtSevenFiveRetRI}{\vtwo{41.8}}  %
\newcommand{\AlphaPtSevenFiveFrzAR}{\vtwo{31.1}}  %
\newcommand{\AlphaPtSevenFiveFrzRI}{\vtwo{10.7}}  %
\newcommand{\AlphaOneRetAR}{\vtwo{83.6}}  %
\newcommand{\AlphaOneRetRI}{\vtwo{40.1}}  %
\newcommand{\AlphaOneFrzAR}{\vtwo{34.3}}  %
\newcommand{\AlphaOneFrzRI}{\vtwo{9.9}}  %
\newcommand{\AlphaTwoRetAR}{\vtwo{83.0}}  %
\newcommand{\AlphaTwoRetRI}{\vtwo{41.1}}  %
\newcommand{\AlphaTwoFrzAR}{\vtwo{28.4}}  %
\newcommand{\AlphaTwoFrzRI}{\vtwo{9.8}}  %
\newcommand{\AlphaRetARmin}{\vtwo{83.0}}  %
\newcommand{\AlphaRetARmax}{\vtwo{84.8}}  %

\newcommand{\WinFourRetAR}{\vtwo{83.8}}  %
\newcommand{\WinFourRetRI}{\vtwo{48.6}}  %
\newcommand{\WinFourFrzAR}{\vtwo{40.0}}  %
\newcommand{\WinFourFrzRI}{\vtwo{10.8}}  %
\newcommand{\WinEightRetAR}{\vtwo{84.0}}  %
\newcommand{\WinEightRetRI}{\vtwo{47.8}}  %
\newcommand{\WinEightFrzAR}{\vtwo{35.3}}  %
\newcommand{\WinEightFrzRI}{\vtwo{10.7}}  %
\newcommand{\WinSixteenRetAR}{\vtwo{83.7}}  %
\newcommand{\WinSixteenRetRI}{\vtwo{46.9}}  %
\newcommand{\WinSixteenFrzAR}{\vtwo{44.7}}  %
\newcommand{\WinSixteenFrzRI}{\vtwo{10.0}}  %
\newcommand{\WinThirtyTwoRetAR}{\vtwo{83.9}}  %
\newcommand{\WinThirtyTwoRetRI}{\vtwo{41.6}}  %
\newcommand{\WinThirtyTwoFrzAR}{\vtwo{37.3}}  %
\newcommand{\WinThirtyTwoFrzRI}{\vtwo{10.0}}  %
\newcommand{\WinSixtyFourRetAR}{\vtwo{83.7}}  %
\newcommand{\WinSixtyFourRetRI}{\vtwo{43.6}}  %
\newcommand{\WinSixtyFourFrzAR}{\vtwo{42.1}}  %
\newcommand{\WinSixtyFourFrzRI}{\vtwo{10.4}}  %
\newcommand{\WinInfRetAR}{\vtwo{83.7}}  %
\newcommand{\WinInfRetRI}{\vtwo{41.9}}  %
\newcommand{\WinInfFrzAR}{\vtwo{29.7}}  %
\newcommand{\WinInfFrzRI}{\vtwo{8.7}}  %

\newcommand{\CrossRawNtuOTAR}{\keep{88.2}}   %
\newcommand{\CrossRawNtuOTRI}{\keep{67.6}}
\newcommand{\CrossRawEtriAR}{\keep{90.8}}
\newcommand{\CrossRawEtriRI}{\keep{45.6}}
\newcommand{\CrossDmrNtuOTAR}{\keep{39.7}}   %
\newcommand{\CrossDmrNtuOTRI}{\keep{13.6}}
\newcommand{\CrossPmrNtuOTAR}{\keep{28.2}}
\newcommand{\CrossPmrNtuOTRI}{\keep{8.4}}
\newcommand{\CrossTmrNtuOTAR}{\keep{24.5}}
\newcommand{\CrossTmrNtuOTRI}{\keep{9.7}}
\newcommand{\CrossTmrEtriAR}{\keep{40.8}}
\newcommand{\CrossTmrEtriRI}{\keep{2.5}}
\newcommand{\CrossNtuOTRetAR}{\vtwo{74.8}}  %
\newcommand{\CrossNtuOTRetRI}{\vtwo{36.5}}  %
\newcommand{\CrossEtriRetAR}{\vtwo{84.9}}  %
\newcommand{\CrossEtriRetRI}{\vtwo{17.8}}  %
\newcommand{\CrossNtuOTFrzAR}{--} %
\newcommand{\CrossNtuOTMixRI}{\vtwo{6.5}}  %
\newcommand{\CrossNtuOTFrzRI}{\vtwo{6.1}}  %
\newcommand{\CrossEtriFrzAR}{\vtwo{39.6}}  %
\newcommand{\CrossEtriFrzRI}{\vtwo{2.0}}  %
\newcommand{\CrossEtriMixAR}{--} %
\newcommand{\CrossEtriMixRI}{\vtwo{1.8}}  %

\newcommand{\PertLegs}{\vtwo{0.251}}  %
\newcommand{\PertSpine}{\vtwo{0.105}}  %
\newcommand{\PertArmsHeadLo}{\vtwo{0.171}}  %
\newcommand{\PertArmsHeadHi}{\vtwo{0.228}}  %
\newcommand{\PertLegSpineRatio}{\vtwo{2.4}}  %

\def\betarowv#1#2#3#4#5#6#7#8{%
  \setres{beta/#1/#2/RetAR}{\vtwo{#3}}\setres{beta/#1/#2/RetRI}{\vtwo{#4}}%
  \setres{beta/#1/#2/FrzAR}{\vtwo{#5}}\setres{beta/#1/#2/FrzRI}{\vtwo{#6}}%
  \setres{beta/#1/#2/MixAR}{\vtwo{#7}}\setres{beta/#1/#2/MixRI}{\vtwo{#8}}}
\betarowv{0.5}{0.1}{88.85}{69.85}{87.00}{63.40}{87.80}{62.50}  %
\betarowv{0.5}{0.2}{87.06}{64.56}{83.00}{47.30}{85.30}{47.00}  %
\betarowv{0.5}{0.3}{85.34}{59.60}{77.70}{33.90}{80.70}{34.80}  %
\betarowv{0.5}{0.5}{83.83}{48.88}{62.80}{19.20}{67.80}{19.70}  %
\betarowv{0.5}{0.7}{84.07}{45.06}{47.30}{12.40}{52.40}{13.00}  %
\betarowv{0.5}{0.9}{83.77}{42.96}{34.60}{9.60}{39.40}{9.40}  %
\betarowv{0.75}{0.1}{88.66}{70.40}{87.80}{67.40}{87.80}{65.00}  %
\betarowv{0.75}{0.2}{86.93}{66.72}{85.00}{53.80}{85.50}{52.90}  %
\betarowv{0.75}{0.3}{85.07}{62.60}{80.80}{41.40}{81.70}{41.30}  %
\betarowv{0.75}{0.5}{83.91}{51.54}{66.20}{24.70}{68.20}{25.00}  %
\betarowv{0.75}{0.7}{83.55}{45.86}{49.50}{16.50}{51.20}{15.80}  %
\betarowv{0.75}{0.9}{83.26}{42.38}{36.50}{11.80}{37.30}{11.00}  %
\betarowv{1.0}{0.1}{88.34}{71.97}{87.70}{66.20}{87.90}{65.30}  %
\betarowv{1.0}{0.2}{87.01}{66.19}{85.00}{52.60}{86.00}{51.70}  %
\betarowv{1.0}{0.3}{85.94}{61.74}{80.70}{40.40}{82.50}{39.20}  %
\betarowv{1.0}{0.5}{84.52}{51.19}{69.00}{24.30}{70.10}{22.60}  %
\betarowv{1.0}{0.7}{83.94}{44.21}{54.00}{15.50}{54.90}{14.30}  %
\betarowv{1.0}{0.9}{83.99}{40.65}{39.80}{11.30}{40.30}{10.00}  %
\betarowv{2.0}{0.1}{88.64}{71.32}{87.70}{68.10}{87.80}{65.50}  %
\betarowv{2.0}{0.2}{86.48}{65.02}{84.60}{53.50}{85.20}{52.60}  %
\betarowv{2.0}{0.3}{85.21}{59.79}{79.20}{40.30}{80.80}{39.90}  %
\betarowv{2.0}{0.5}{83.52}{50.55}{64.90}{23.30}{65.80}{22.80}  %
\betarowv{2.0}{0.7}{83.65}{46.02}{47.70}{14.60}{49.50}{13.90}  %
\betarowv{2.0}{0.9}{83.23}{43.15}{33.70}{10.90}{35.70}{9.80}  %
\betarowv{train}{0.2}{84.56}{46.87}{59.80}{12.50}{59.80}{10.90}  %
\betarowv{train}{0.3}{84.14}{48.33}{70.60}{10.20}{68.50}{9.50}  %
\betarowv{train}{0.5}{83.79}{43.38}{44.30}{10.00}{44.40}{8.70}  %
\betarowv{train}{0.7}{84.06}{44.92}{47.70}{11.40}{51.60}{9.50}  %

\newcommand{\BtrainNumSeeds}{\vtwo{5}}
\newcommand{\BtrainFrzAR}{\vtwo{61.1}}      %
\newcommand{\BtrainFrzARsd}{\vtwo{5.7}}     %
\newcommand{\BtrainRetAR}{\vtwo{84.1}}      %
\newcommand{\BtrainRetARsd}{\vtwo{0.4}}     %
\newcommand{\BtrainRetRI}{\vtwo{48.7}}      %
\newcommand{\BtrainRetRIsd}{\vtwo{2.6}}     %
\newcommand{\BtrainAdaptRI}{\vtwo{48.8}}    %
\newcommand{\BtrainAdaptRIsd}{\vtwo{2.8}}   %

%% file: sec_intro.tex
\section{Introduction}
\label{sec:intro}

Skeleton sequences, the 3D positions of a person's joints over time, are the working format of action recognition~\citep{shahroudy2016ntu}, elderly-care robotics~\citep{jang2020etri}, physical rehabilitation~\citep{capecci2019kimore,vakanski2018uiprmd}, and immersive reality~\citep{carr2024metacom,lin2022digital}. Dropping appearance does not remove identity. Joint coordinates alone support re-identification (about 80\% top-1 over the 40 NTU subjects~\citep{Moon2023SkeletonAnon}) and cross-clip linkage~\citep{carr2023linkage}; head-and-hand telemetry identified 55,541 VR users with 94\% accuracy from 100\,s of motion~\citep{nair2023unique}; and VR sensor data supports user identification across applications~\citep{jarin2025behavr}. Dedicated gait models recognize people from skeleton sequences~\citep{teepe2021gaitgraph,teepe2022gaitgraph2,gaitMixer}. Once a capture device ships skeleton streams to a third-party analytics service, that service, or anyone who obtains the stream, can attempt to link the records to the people who produced them.

A curator who wants to share such streams for analytics faces three deployment constraints (Section~\ref{sec:threat-req} adds a fourth, bounded state). The consumers usually run action classifiers that were trained on raw skeletons and cannot be retrained for each privacy setting, so the de-identified output must stay in the raw coordinate space (\emph{drop-in}). The stream arrives frame by frame, so the transformation must use only past and present frames (\emph{causal}). And the curator has no matching reference person for every capture, so the transformation must not need one (\emph{target-free}). Adversarial anonymizers train a transformation against an identity classifier, first for video~\citep{wu2018privacy} and then for skeletons~\citep{Moon2023SkeletonAnon}. Moon et al.'s residual variant adds a per-frame MLP correction in the raw coordinate space while a frozen action classifier keeps actions recognizable, so it meets all three constraints by design; because it maps each frame independently, however, it uses no temporal context, and it was evaluated against pre-trained classifiers rather than attackers that train on its output. Motion retargeting maps motion onto a target skeleton~\citep{aberman2020character,carr2025privacy,carr2026preserving}: it needs that target at inference, its strongest variant processes complete sequences, and its simpler variants move the output so far from the raw distribution that raw-trained classifiers fail on it (Section~\ref{sec:res-frontier}).

We present \mir{},\footnote{The acronym is unrelated to other systems named Mirage in computer science, such as the MirageOS library operating system; we use it only for the anonymizer described here. ``Privacy-preserving'' in the title names the goal, not a guarantee.} which keeps the raw-coordinate residual and adds temporal context: a causal encoder--decoder Transformer computes a residual $\boldsymbol\delta_t$ for each incoming frame by attending over the frames seen so far, and releases $\hat{\mathbf s}_t=\mathbf s_t+\beta\boldsymbol\delta_t$. Attention over past frames lets the correction depend on how a person moves, not only on the pose in one frame. The residual keeps the output in the input's coordinate space, and adversarial identity heads, cooperative action heads, a distribution discriminator and an adversarial contrastive probe shape it during training. One causal, optionally windowed attention mask applies to encoder self-attention, decoder self-attention and decoder cross-attention; positions enter through rotary embeddings; and the decoder is conditioned on the previous \emph{source} frame, which is available online. As a result, the batched computation used for training and the frame-by-frame computation used at deployment are the same function, and with a finite window per-stream state is bounded; our main results use an unbounded window, whose state grows with stream length (Section~\ref{sec:method}).

\mir{} shares being target-free and causal with the per-frame residual anonymizer of Moon et al.; what it adds is temporal context and an evaluation against adaptive and identity-specialized attackers. We train Moon et al.'s anonymizer under our protocol and report it next to \mir{} (Tables~\ref{tab:frontier} and~\ref{tab:retrained}). We judge both adaptive-first. When an identity classifier is retrained on each method's output, SGN identifies \RetSgnRI\% of test clips on \mir{} output against \MoonRetSgnRI\% on Moon et al.'s, while a consumer who retrains an action classifier recovers \RetSgnAR\% and \MoonRetSgnAR\% respectively (raw \RawSgnAR\%). This retrained comparison is our main result. Retargeting leaks less under retraining (\TmrRetSgnRI\% retrained SGN RI for DisentangledTMR) but needs a target skeleton and whole clips, and retrained action recognition on its output reaches only \TmrRetSgnAR\%.

\mir{} does not meet the drop-in constraint above. Raw-trained (frozen) classifiers recognize only \MainSgnAR\% of actions on its output at the default residual scale, against \TmrSgnAR\% for the strongest retargeting baseline, DisentangledTMR~\citep{carr2026preserving}; frozen re-identification is also lower (\MainSgnRI\% vs.\ \TmrSgnRI\%), and all four frozen differences are significant over five seeds per method (Table~\ref{tab:sig}). The gap between frozen and retrained action accuracy shows that the output is shifted away from the distribution raw-trained classifiers expect; shrinking the residual at inference recovers frozen accuracy only by giving back privacy (Section~\ref{sec:res-ops}). The strongest adaptive attacker, \WorstRetAttacker{}, still identifies \WorstRetRI\% of clips (chance \ChanceRI\%), and identification risk is unevenly spread across individuals (Section~\ref{sec:res-suite}). The protection is empirical identity-leakage reduction; it carries no formal guarantee.

\textbf{Contributions.}
\begin{itemize}
  \item A system and threat model for releasing de-identified skeleton streams to consumers that run raw-trained models, with a passive (A1) and an adaptive (A2) adversary for closed-set re-identification and explicit non-goals (Section~\ref{sec:threat}).
  \item \mir{}, a target-free, causal anonymizer with output in the raw coordinate space that extends per-frame residual anonymization~\citep{Moon2023SkeletonAnon} with causal attention over past frames, and whose training computation equals its online computation, with state bounded under a finite window and positions that do not stop at the training length; we verify batch--stream equality on all \EqNumSeq{} NTU~60 training and test sequences and bounded memory on a stream far longer than any training clip (Sections~\ref{sec:method} and~\ref{sec:res-stream}).
  \item An evaluation against both adversaries with ten frozen action-recognition families, two identity-specialized gait models (frozen and retrained), retrained and adaptive classifiers, 1-NN embedding matching, and our published linkage attack; a per-frame residual baseline trained under the same protocol; per-subject tail analysis; and Welch tests over generator seeds against the retargeting baseline's seed distribution (Sections~\ref{sec:eval}--\ref{sec:results}).
\end{itemize}

%% file: sec_related.tex
\section{Background and Related Work}
\label{sec:related}

\textbf{Identity in skeleton motion.} Skeletons carry static anthropometrics (limb lengths, proportions) and dynamic style (gait, timing, posture). Both support re-identification from joint coordinates~\citep{Moon2023SkeletonAnon,miller2020personal,nair2023unique,jarin2025behavr}. Our earlier work~\citep{carr2023linkage} showed that a pairwise verifier links skeleton recordings of the same person across clips, which we reuse as an attacker. Skeleton gait recognizers, GaitGraph~\citep{teepe2021gaitgraph}, its successor GaitGraph2~\citep{teepe2022gaitgraph2}, and GaitMixer~\citep{gaitMixer}, are built for identification rather than for actions, which makes them the natural identity-specialized attackers. Hanisch et al.~\citep{hanisch2024false} show that anonymization studies for faces and gait often assume a weak adversary that is unaware of the anonymization, and that this overstates protection; they argue for a strong, anonymization-aware adversary. Our adaptive adversary A2 follows that recommendation.

\textbf{Action recognition as the utility consumer.} Skeleton action recognition is dominated by graph convolutional networks such as ST-GCN~\citep{YanXL18}, 2s-AGCN~\citep{shi2019twoStream}, CTR-GCN~\citep{chen2021ctrgcn}, MS-G3D~\citep{liu2020disentangling}, HD-GCN~\citep{lee2023hdgcn} and InfoGCN~\citep{chi2022infogcn}, the semantics-guided SGN~\citep{zhang2020semantics}, and Transformers such as Skeleton-MixFormer~\citep{xin2023skeleton}, FreqMixFormer~\citep{wu2024frequency} and SkateFormer~\citep{do2024skateformer}. We treat these models, trained on raw data, as the deployed consumers whose behavior a drop-in anonymizer must preserve, and we also repurpose them as identity classifiers.

\textbf{Adversarial anonymization.} These methods keep the input format and train a transformation against an identity adversary while preserving the task~\citep{wu2018privacy,Moon2023SkeletonAnon}, using min--max or gradient-reversal training~\citep{ganin2016dann}; a related line learns privacy-preserving latent representations instead~\citep{xiao2020adversarial}. The closest prior work is Moon et al.~\citep{Moon2023SkeletonAnon}. Their residual anonymizer $f(\mathbf x)=\mathrm{MLP}(\mathbf x)+\mathbf x$ is applied to each frame independently, outputs raw coordinates and needs no target; it is trained against an identity classifier co-trained as its adversary and an action classifier that stays frozen, and it is evaluated with pre-trained classifiers other than those used in training. They also propose a U-Net anonymizer that is not frame-wise. The residual variant therefore meets R1, R2 and R4 of Section~\ref{sec:threat-req} by construction and R3 in its authors' evaluation, but its correction for frame $t$ sees only frame $t$. \mir{} keeps the residual formulation and adds causal attention over past frames, and we train Moon et al.'s residual anonymizer under our protocol as a baseline. A recognized difficulty for such methods is protection beyond the adversaries used during optimization: Wu et al. restart and ensemble the adversary to generalize to unseen models~\citep{wu2018privacy}, and Moon et al. find that adversarial attacks on one identity model leave a different model able to identify~\citep{Moon2023SkeletonAnon}. Protection measured against pre-trained classifiers can therefore overstate what an attacker who trains on the output recovers, which is why we evaluate both passive and adaptive attackers from many families.

\textbf{Motion retargeting for privacy.} Retargeting transfers the motion of a source performer onto the skeleton of a target performer~\citep{villegas2018neural,aberman2020character}. Applied to privacy, it replaces the source's body proportions, the strongest static identity cue, with the target's. We compare against three methods from this line, all from our own earlier work and all needing a target skeleton at inference.
\emph{DMR}~\citep{carr2025privacy} adapts deep motion retargeting~\citep{aberman2020character} to NTU skeletons: separate encoders extract a motion representation from the source and a static skeleton representation from the target, and a decoder combines them, with no privacy-specific training.
\emph{PMR}~\citep{carr2025privacy} adds cooperative and adversarial identity and action classifiers on the motion and body embeddings of the same pipeline, so that the motion code carries no actor identity, plus a realism discriminator on the output.
\emph{DisentangledTMR}~\citep{carr2026preserving} replaces the convolutional retargeting network with Transformers that factorize motion and body representations; a Transformer decoder generates the retargeted sequence autoregressively from the source's motion code and a target's body code, is trained in three stages, and has \CostTmrParams{}M parameters. Although its decoder is autoregressive, its encoders consume the complete source clip, so it is not causal end to end and runs offline. Among the retargeting methods it has the highest frozen-evaluator AR, and we use it as the main baseline, in its $\beta{=}0.2$ configuration (its own loss weight, unrelated to our residual scale $\beta$) with five training seeds.
The convolutional retargeting methods place their output in a skeleton space that raw-trained classifiers do not recognize (frozen AR of \DmrSgnAR\% and \PmrSgnAR\% for DMR and PMR, Table~\ref{tab:frontier}), while DisentangledTMR keeps frozen AR high.

\textbf{Other motion anonymizers.} Deep motion masking~\citep{nair2024deep} anonymizes VR head-and-hand telemetry in real time and was evaluated in a user study. Pantomime~\citep{hanisch2026pantomime} re-synthesizes full-body motion with foundation motion models so that the result stays natural, and reports identification accuracy reduced to 10\%. These works target different data (VR telemetry) or different goals (naturalness of re-synthesized motion) than compatibility with deployed skeleton action recognizers; we have not evaluated them under our protocol, and a head-to-head comparison is future work.

\textbf{Causal Transformers.} \mir{} uses the Transformer~\citep{vaswani2017attention} with causal masks, optional local windows as in sparse attention~\citep{child2019generating}, and rotary position embeddings~\citep{su2024roformer}, through which position enters each attention score only via the relative offset. Transformer motion generators, autoregressive~\citep{zhao2024dartcontrol,zhang2023t2mgpt} or whole-sequence~\citep{xu2023actformer}, use similar building blocks for synthesis rather than privacy. The training objective draws on supervised contrastive learning~\citep{khosla2020supervised} and adversarial distribution matching~\citep{goodfellow2014generative}.

%% file: sec_threat.tex
\section{System and Threat Model}
\label{sec:threat}

\subsection{Setting}
\label{sec:threat-setting}

A \emph{curator}, for example a VR platform, a clinic running gait assessments, or an ambient-sensing deployment, captures skeleton streams from \emph{data subjects} and passes them to \emph{consumers}: analytics services, research partners, or vendor APIs that run action classifiers. The curator applies an anonymizer $M$ at its boundary, on the capture device or an edge server, and releases only $\hat{\mathbf S}=M(\mathbf S)$. Consumers run their existing action classifiers, which were trained on raw skeletons, on $\hat{\mathbf S}$ without modification; some consumers may also retrain models on released data. Following Kerckhoffs's principle, we assume the anonymizer's architecture and trained weights may be public, for example because the curator ships the anonymizer in a device or publishes it. \mir{} is deterministic: the same input always yields the same output.

\subsection{Requirements}
\label{sec:threat-req}

A deployable anonymizer for this setting must satisfy four requirements; Table~\ref{tab:constraints} shows which methods meet them.
\textbf{R1 Target-free:} $M$ takes no reference skeleton or other auxiliary person.
\textbf{R2 Causal:} $\hat{\mathbf s}_t$ is a function of $\mathbf s_{1:t}$ only, so it can be emitted when frame $t$ arrives.
\textbf{R3 Drop-in:} $\hat{\mathbf s}_t$ is in the coordinate space of $\mathbf s_t$ and raw-trained classifiers keep working on it, which we measure as frozen AR.
\textbf{R4 Bounded state:} per-stream memory does not grow with stream length, at least for a configuration the deployer can choose.

\begin{table}[t]
\centering
\small
\caption{Deployment requirements (Section~\ref{sec:threat-req}). R3 is judged by frozen action recognition in Table~\ref{tab:frontier}: DMR and PMR fail it, DisentangledTMR meets it empirically although its output carries the target's body proportions, and \mir{} fails it at $\beta{=}1$ (frozen SGN AR \MainSgnAR\%). The retargeting methods process whole clips offline, so R4 does not apply. Context: input frames the output at time $t$ can depend on. $^\S$Residual variant; R3 by design and in its authors' evaluation, and under our protocol (Table~\ref{tab:frontier}). $^\ddagger$Raw coordinate space, but raw-trained classifiers lose most of their accuracy at $\beta{=}1$; lowering $\beta$ at inference trades privacy back for frozen accuracy (Section~\ref{sec:res-ops}). $^\ast$\mir{}'s state is bounded for a finite window $W$; with $W{=}\infty$ it grows linearly.}
\label{tab:constraints}
\setlength{\tabcolsep}{3pt}
\begin{tabular}{l ccccc}
\toprule
& R1 & R2 & R3 & R4 & \\
Method & target-free & causal & drop-in & bounded & context \\
\midrule
Gaussian noise & \cmark & \cmark & \xmark & \cmark & frame \\
DMR, PMR~\citep{carr2025privacy} & \xmark & \xmark & \xmark & n/a & clip \\
DisentangledTMR~\citep{carr2026preserving} & \xmark & \xmark & \cmark & n/a & clip \\
Moon et al.~\citep{Moon2023SkeletonAnon} & \cmark & \cmark & \cmark$^\S$ & \cmark & frame \\
\mir{} (ours) & \cmark & \cmark & \xmark$^\ddagger$ & \cmark$^\ast$ & past \\
\bottomrule
\end{tabular}
\end{table}

\subsection{Adversary}
\label{sec:threat-adv}

\textbf{Goal.} Closed-set re-identification: given a released sequence $\hat{\mathbf S}$, decide which of $K$ enrolled people produced it. In our benchmark the gallery is the $K{=}\NumSubjects$ subjects of NTU RGB+D~60, so random guessing succeeds with probability \ChanceRI\%. We report top-1 accuracy (RI) on held-out sequences of gallery members.

\textbf{Benefit.} Re-identification links behavioral records to a person: from motion data alone, an adversary can identify a VR user and track that person across sessions or apps~\citep{nair2023unique,jarin2025behavr}. By extension, the same capability would let an adversary attach a gait assessment or workplace activity logs to a known individual, or join such records with other data about them.

\textbf{Prior knowledge.} The adversary holds labeled raw recordings of each gallery member, for example from an earlier enrollment, a leaked dataset, or a different camera view. In our protocol these are the training-split recordings of the same \NumSubjects{} subjects, captured from different camera views than the test recordings.

\textbf{Capabilities.} We consider two adversaries.
\emph{A1, passive.} A1 trains identity models on its labeled raw recordings and applies them unchanged to released data. It does not know or use the anonymizer. This is the threat posed by off-the-shelf identification models and by consumers who already hold raw-trained identity models. We instantiate A1 with ten action-recognition backbones repurposed as identity classifiers, two gait recognizers, 1-NN matching of released clips against raw gallery embeddings, and our pairwise linkage verifier~\citep{carr2023linkage}.
\emph{A2, adaptive.} A2 additionally holds the released anonymizer, either its weights or query access. It runs its labeled raw recordings through the anonymizer to obtain labeled anonymized data and trains identity models on that distribution, so it is aware of the anonymization as recommended by Hanisch et al.~\citep{hanisch2024false}. We instantiate A2 with SGN and MixFormer trained from scratch on anonymized data for 60 epochs (\emph{retrained}) and for 120 epochs with identity-specific tuning (\emph{adaptive}), and with GaitGraph2 and GaitMixer trained on anonymized data. A learned inversion network followed by a raw identity model is a special case of these end-to-end attackers; we do not evaluate per-sample gradient-based inversion of the released weights.

A1 is the relevant model when the anonymizer is kept private or when the attacker reuses existing identification models; A2 is the conservative model for a public anonymizer. We report both and treat A2 as the headline privacy number (abstract, Section~\ref{sec:res-adaptive}).

\textbf{Defender utility.} The defender wants consumers who retrain on released data to reach high action accuracy (retrained AR) and, for drop-in use, raw-trained consumers to keep recognizing actions (frozen AR). We report both; \mir{} at $\beta{=}1$ serves only the first (Section~\ref{sec:res-frontier}).

\subsection{Non-Goals}
\label{sec:threat-nongoals}

\textbf{Terminology.} We follow the literature in calling such transformations \emph{anonymizers}; the word names the intent. What we measure is de-identification: reduced identification accuracy against specific attackers.

\textbf{No formal guarantee.} \mir{} provides empirical identity-leakage reduction against the attackers we evaluate. It offers no differential privacy, $k$-anonymity, or other provable bound, and a stronger or differently informed attacker may recover more identity. Training-participation privacy for the people whose motion trains an anonymizer is a different property that we do not address.

\textbf{Out of scope.} Open-set identification, verification against people outside the gallery, and linkage of one person's clips or sessions to each other are examined only through the linkage diagnostic of Section~\ref{sec:res-suite}, not as full attack settings. Attribute inference (for example sex, age, or health conditions), multi-person scenes, and adversaries who see the raw stream or control the curator are out of scope. The released stream reveals that it was anonymized; we do not hide that fact.

%% file: sec_method.tex
\section{MIRAGE}
\label{sec:method}

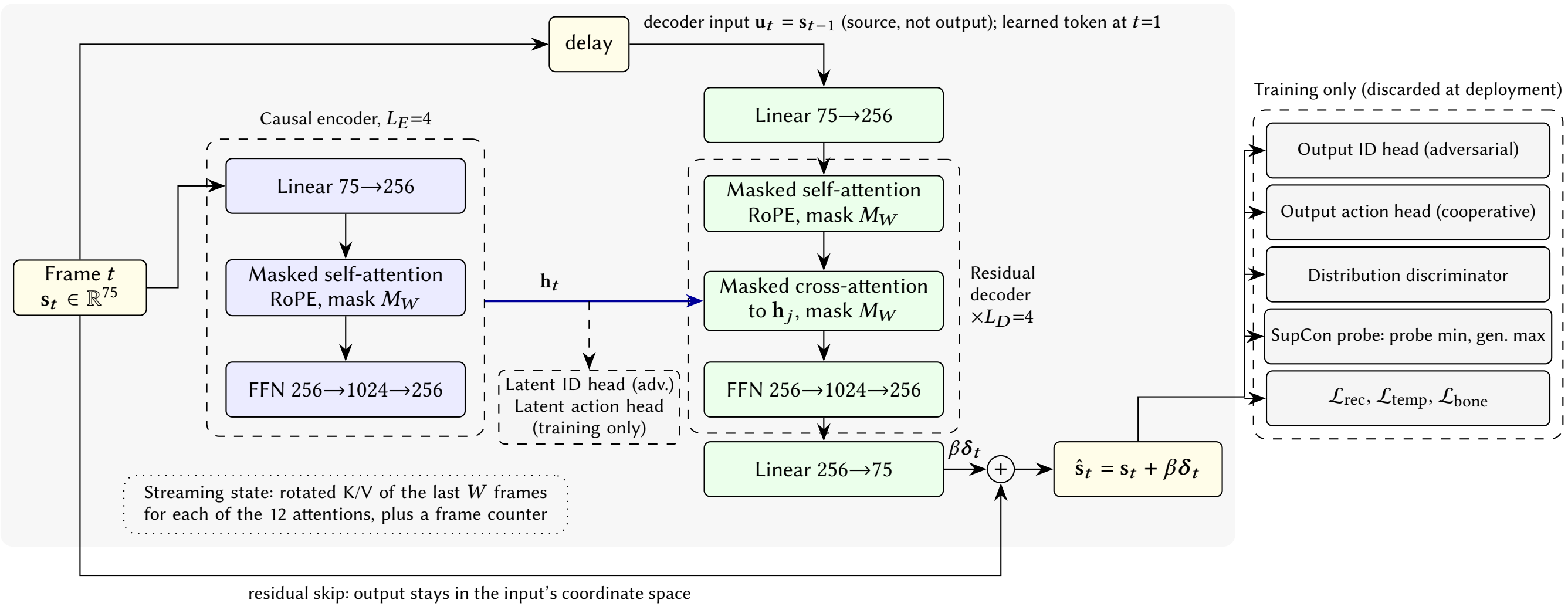
\begin{figure*}[t]
\centering
\resizebox{\textwidth}{!}{\input{figures/fig_architecture}}
\caption{\mir{} at deployment (shaded) and in training. Every attention, in both encoder and decoder, uses the same causal mask $M_W$ (query $t$ sees positions $t-W<j\le t$) and rotary positions. The decoder's input at step $t$ is the previous source frame, so the batched training computation and the frame-by-frame computation coincide, and the streaming state is the last $W$ keys and values of each attention. Heads, discriminator and contrastive probe are used only in training.}
\Description{Block diagram. On the left, an input frame s_t enters a causal encoder of four layers, each with a linear input projection, masked self-attention with rotary positions, and a feed-forward network. The encoder output h_t feeds a residual decoder of four layers with masked self-attention, masked cross-attention to the encoder states, and a feed-forward network. The decoder input is the previous source frame obtained through a delay from the input. The decoder output, scaled by beta, is added to the input frame to give the anonymized frame. A note marks the streaming state as the last W keys and values of each of the twelve attentions. On the right, a training-only group contains an output identity head, an output action head, a distribution discriminator, a supervised contrastive probe, and motion regularizers; latent identity and action heads read the encoder output.}
\label{fig:architecture}
\end{figure*}

\subsection{Residual Formulation}
\label{sec:method-residual}

A frame $\mathbf s_t\in\mathbb R^{JC}$ holds $J{=}25$ joints with $C{=}3$ coordinates. \mir{} emits
\begin{equation}
\label{eq:residual}
\hat{\mathbf s}_t=\mathbf s_t+\beta\,\boldsymbol\delta_t,\qquad \boldsymbol\delta_t=G_\theta(\mathbf s_{1:t}),
\end{equation}
where $G_\theta$ is the encoder--decoder generator of Figure~\ref{fig:architecture}, $\beta{=}1$ during training, and $\beta\in(0,1]$ can be lowered at inference to attenuate the perturbation without retraining (\emph{\mir{}-$\beta$}). Because the output is the input plus a learned correction, it lives in the raw coordinate space by construction; whether raw-trained classifiers still read it correctly is what the distribution-matching losses below are for, and is measured, not assumed.

\subsection{One Causal Mask for Every Attention}
\label{sec:method-causal}

Let $W\in\{1,2,\dots\}\cup\{\infty\}$ be the context window and define the mask
\begin{equation}
\label{eq:mask}
M_W(t,j)=1 \iff t-W<j\le t .
\end{equation}
\textbf{Encoder.} A linear map embeds $\mathbf s_t$ in $\mathbb R^{d}$, $d{=}256$, and $L_E{=}4$ pre-norm layers apply masked multi-head self-attention (8 heads) and a feed-forward network ($256{\to}1024{\to}256$, GELU); a final layer norm gives $\mathbf h_t$.
\textbf{Decoder.} The decoder input at step $t$ is the previous source frame, $\mathbf u_t=\mathbf s_{t-1}$, with a learned token $\mathbf u_1=\mathbf e$. A linear map embeds $\mathbf u_t$, and $L_D{=}4$ pre-norm layers apply masked self-attention over the embedded inputs, masked cross-attention whose queries are the decoder states and whose keys and values are the encoder states $\mathbf h_j$, and a feed-forward network; a final linear map to $\mathbb R^{JC}$ gives $\boldsymbol\delta_t$.
\textbf{Positions.} Every attention rotates queries and keys by their absolute frame index with rotary embeddings~\citep{su2024roformer}, so position enters each attention score only through the offset $j-t$. There is no learned position table, hence no maximum sequence length.

All three kinds of attention, encoder self, decoder self and decoder cross, use the same mask $M_W$. This gives the property the online claim rests on.

\textbf{Property (causality and receptive field).} \emph{For every $t$, $\hat{\mathbf s}_t$ depends only on $\mathbf s_{1:t}$. For finite $W\ge2$ it depends only on the last $R=(L_E+L_D)(W-1)+1$ frames.} Each masked attention at position $t$ reads positions $j\le t$ of the layer below, the encoder's input at $j$ is $\mathbf s_j$, and the decoder's input at $j$ is $\mathbf s_{j-1}$; induction over layers gives causality. Each layer can reach $W-1$ frames further back, and the deepest path runs through $L_D-1$ decoder self-attentions, one cross-attention and $L_E$ encoder self-attentions, which gives $R$. For the windowed configuration $W{=}16$ this is 121 frames.

\subsection{Streaming with Bounded State}
\label{sec:method-stream}

At deployment \mir{} processes one frame at a time. Its state is the frame counter $t$, the previous source frame, and, for each of the 12 attentions (4 encoder self, 4 decoder self, 4 decoder cross), the rotated keys and the values of the last $W$ positions. When $\mathbf s_t$ arrives, each layer projects only the new position, appends its key and value to that attention's cache, drops entries older than $W$, and attends from the new query to the cache; cross-attention caches the projected encoder states $\mathbf h_j$. The masks, positions and inputs are exactly those of the batched computation, so batched and frame-by-frame outputs agree up to floating-point error. Unit tests pin this equality, prefix invariance (changing frames after $t$ does not change outputs up to $t$), padding independence, and bounded caches on streams longer than the training clips; Section~\ref{sec:res-stream} measures it on all \EqNumSeq{} NTU~60 training and test sequences.

The cache holds $12\times2\times W\times d$ values, 24\,KiB per cached frame in 32-bit floats, so $W{=}16$ needs 384\,KiB per stream. Per-frame attention cost is $O((L_E+2L_D)\,W d)$ on top of the fixed projection and feed-forward cost. With $W{=}\infty$, the configuration of our main results, both grow linearly with stream length, so R4 holds only for finite $W$. Rotation angles are computed in double precision from the frame counter, so positions keep advancing after the cache fills and streams can run far past the 300-frame clips seen in training.

\textbf{Sense of ``autoregressive''.} \mir{} is autoregressive over its input stream: each output is generated from the history of source frames through a decoder whose inputs are earlier frames. It does not feed its own outputs back. Conditioning on $\mathbf s_{t-1}$ rather than $\hat{\mathbf s}_{t-1}$ makes the teacher-forced training input identical to the inference input, so there is no exposure bias and no mismatch between the trained and deployed function.

\subsection{Training Objective}
\label{sec:method-train}

Training alternates a classifier step and a generator step on mini-batches of whole clips, computed with the batched path.

\textbf{Supervision heads.} Two \emph{latent heads} read the encoder states (masked temporal mean, then a two-layer MLP) and predict identity and action; two \emph{output heads} read $\hat{\mathbf S}$ (three 1D convolutions, temporal pooling, two-layer MLP) and do the same. The generator is adversarial to both identity heads and cooperative with both action heads.

\textbf{Distribution discriminator.} A temporal CNN with spectral normalization scores whole sequences as raw or anonymized. The generator is trained to make its output indistinguishable from raw data. Unlike the identity heads, the discriminator supervises distributional statistics rather than a label; removing it lowers frozen action accuracy further (Section~\ref{sec:res-ops}).

\textbf{Adversarial contrastive probe.} A probe $p_\phi$ maps the masked temporal mean of $\hat{\mathbf S}$ through a two-layer MLP ($75{\to}128{\to}128$) to a unit vector $\mathbf z$. In the classifier step the probe \emph{minimizes} a supervised contrastive loss~\citep{khosla2020supervised} over identity labels,
\begin{equation}
\label{eq:supcon}
\mathcal L_{\mathrm{con}}=-\frac{1}{|\mathcal P|}\sum_{(i,j)\in\mathcal P}\log\frac{\exp(\mathbf z_i^\top\mathbf z_j/\tau)}{\sum_{k\neq i}\exp(\mathbf z_i^\top\mathbf z_k/\tau)},
\end{equation}
with $\mathcal P$ the same-identity pairs in the batch and $\tau{=}0.1$, so it learns to cluster recordings of the same person; in the generator step the generator \emph{maximizes} $\mathcal L_{\mathrm{con}}$ against the current probe. The probe is therefore an adversary that the generator must scatter, not a module that can scatter identities on its own.

\textbf{Objective.} With generator parameters $\theta$, heads $\psi$, discriminator $\omega$ and probe $\phi$, the classifier step minimizes the four heads' cross-entropies, the discriminator's binary cross-entropy, and $\mathcal L_{\mathrm{con}}$ over $(\psi,\omega,\phi)$ on detached generator outputs. The generator step minimizes
\begin{equation}
\label{eq:gen}
\begin{aligned}
\mathcal L_G={}&-\alpha\,w_e\!\!\sum_{\ell\in\{\mathrm{lat},\mathrm{out}\}}\!\!\mathcal L^{\ell}_{\mathrm{id}}+\!\!\sum_{\ell\in\{\mathrm{lat},\mathrm{out}\}}\!\!\mathcal L^{\ell}_{\mathrm{act}}-\lambda_{\mathrm{con}}\mathcal L_{\mathrm{con}}+\lambda_{\mathrm{dd}}\mathcal L^{G}_{\mathrm{dd}}\\
&+\lambda_{\mathrm{rec}}\mathcal L_{\mathrm{rec}}+\lambda_{\mathrm{temp}}\mathcal L_{\mathrm{temp}}+\lambda_{\mathrm{bone}}\mathcal L_{\mathrm{bone}},
\end{aligned}
\end{equation}
where $\mathcal L_{\mathrm{rec}}$, $\mathcal L_{\mathrm{temp}}$ and $\mathcal L_{\mathrm{bone}}$ penalize squared deviations of positions, frame-to-frame velocities, and bone lengths from the input (Appendix~\ref{app:impl}). The privacy weight $\alpha$ is the only weight we tune; $\lambda_{\mathrm{con}}{=}\lambda_{\mathrm{dd}}{=}\lambda_{\mathrm{rec}}{=}1$ and $\lambda_{\mathrm{temp}}{=}\lambda_{\mathrm{bone}}{=}0.5$. The adversarial identity term is ramped in by $w_e=\min(1,e/E_w)$ over the first $E_w{=}10$ epochs so that the action heads shape the output before identity pressure starts.

\textbf{Deployment footprint.} Only the generator is kept: \CostMirageParams{}M parameters (encoder, decoder, start token). Heads, discriminator and probe are discarded.

%% file: figures/fig_architecture.tex
\begin{tikzpicture}[
  font=\sffamily\footnotesize,
  >={Stealth[length=2mm]},
  blk/.style={draw, rounded corners=2pt, align=center, minimum width=2.7cm, minimum height=0.62cm, inner sep=2pt},
  enc/.style={blk, fill=blue!8},
  dec/.style={blk, fill=green!8},
  head/.style={blk, minimum width=3.2cm, fill=gray!8, font=\sffamily\scriptsize},
  note/.style={align=center, font=\sffamily\scriptsize},
]
\node[enc] (e1) at (3.6, 1.15) {Linear $75{\to}256$};
\node[enc] (e2) at (3.6, 0.0) {Masked self-attention\\RoPE, mask $M_W$};
\node[enc] (e3) at (3.6,-1.15) {FFN $256{\to}1024{\to}256$};
\draw[->] (e1) -- (e2); \draw[->] (e2) -- (e3);
\node[draw, dashed, rounded corners, fit=(e1)(e3), inner sep=6pt, label={[font=\sffamily\scriptsize]above:Causal encoder, $L_E{=}4$}] (encbox) {};
\node[dec] (d0) at (9.0, 1.95) {Linear $75{\to}256$};
\node[dec] (d1) at (9.0, 0.95) {Masked self-attention\\RoPE, mask $M_W$};
\node[dec] (d2) at (9.0,-0.15) {Masked cross-attention\\to $\mathbf h_j$, mask $M_W$};
\node[dec] (d3) at (9.0,-1.15) {FFN $256{\to}1024{\to}256$};
\node[dec] (d4) at (9.0,-2.05) {Linear $256{\to}75$};
\draw[->] (d0) -- (d1); \draw[->] (d1) -- (d2); \draw[->] (d2) -- (d3); \draw[->] (d3) -- (d4);
\node[draw, dashed, rounded corners, fit=(d1)(d3), inner sep=5pt] (decbox) {};
\node[note, anchor=west, align=left] at (decbox.east) {Residual\\decoder\\$\times L_D{=}4$};
\node[blk, minimum width=1.5cm, fill=yellow!10] (in) at (0.6, 0.0) {Frame $t$\\$\mathbf s_t\in\mathbb R^{75}$};
\draw[->] (in.east) -- ++(0.35,0) |- (e1.west);
\node[blk, minimum width=0.9cm, fill=yellow!10] (delay) at (6.35, 2.75) {delay};
\draw[->] (in.north) |- (delay.west);
\draw[->] (delay.east) -| (d0.north);
\node[note, anchor=south west] at (6.85, 2.78) {decoder input $\mathbf u_t=\mathbf s_{t-1}$ (source, not output); learned token at $t{=}1$};
\draw[->, thick, blue!60!black] (encbox.east |- d2.west) -- (d2.west)
  node[pos=0.3, above, note, text=black] {$\mathbf h_t$};
\node[head, dashed, minimum width=2.0cm, text width=1.9cm] (lat) at (6.35,-1.35) {Latent ID head (adv.)\\Latent action head\\(training only)};
\draw[->, dashed] (6.35,-0.15) -- (lat.north);
\node[draw, circle, inner sep=1pt] (plus) at (11.0,-2.05) {$+$};
\draw[->] (d4.east) -- (plus.west) node[midway, above, note] {$\beta\boldsymbol\delta_t$};
\draw[->] (in.south) |- (0.6,-3.25) -| (plus.south);
\node[note, anchor=north] at (5.0,-3.25) {residual skip: output stays in the input's coordinate space};
\node[blk, minimum width=1.9cm, fill=yellow!10] (out) at (12.55,-2.05) {$\hat{\mathbf s}_t=\mathbf s_t+\beta\boldsymbol\delta_t$};
\draw[->] (plus.east) -- (out.west);
\node[draw, dotted, rounded corners, note, text width=4.8cm] (cache) at (3.6,-2.45)
  {Streaming state: rotated K/V of the last $W$ frames for each of the 12 attentions, plus a frame counter};
\node[head] (h1) at (15.6, 1.55) {Output ID head (adversarial)};
\node[head] (h2) at (15.6, 0.85) {Output action head (cooperative)};
\node[head] (h3) at (15.6, 0.15) {Distribution discriminator};
\node[head] (h4) at (15.6,-0.55) {SupCon probe: probe min, gen.\ max};
\node[head] (h5) at (15.6,-1.25) {$\mathcal L_{\mathrm{rec}},\mathcal L_{\mathrm{temp}},\mathcal L_{\mathrm{bone}}$};
\draw[->] (out.north) -- ++(0,0.5) -| (13.75,1.55) -- (h1.west);
\foreach \n in {h2,h3,h4,h5} { \draw[->] (13.75,0 |- \n.west) -- (\n.west); }
\node[draw, dashed, rounded corners, fit=(h1)(h5), inner sep=4pt, label={[note]above:Training only (discarded at deployment)}] {};
\begin{scope}[on background layer]
  \node[fill=gray!6, rounded corners, fit=(in)(encbox)(decbox)(d0)(d4)(delay)(out)(cache), inner sep=4pt] {};
\end{scope}
\end{tikzpicture}

%% file: sec_eval.tex
\section{Evaluation Methodology}
\label{sec:eval}

\textbf{Data.} The primary benchmark is NTU RGB+D~60~\citep{shahroudy2016ntu}, Kinect~v2 skeletons of \NumSubjects{} subjects, with the 49 single-person actions and the cross-view (X-View) split: training and test recordings come from different cameras, and every subject appears in both. RI is \NumSubjects-way closed-set classification of the subject (chance \ChanceRI\%); AR is 49-way action classification (chance about \ChanceAR\%). Action scores use the 15,474 single-person-action test clips; the primary SGN and MixFormer RI uses all 18,932 X-View test clips, including two-person actions, with one skeleton per clip (Table~\ref{tab:suite}). Cross-dataset runs on NTU RGB+D~120~\citep{liu2019ntu} (106 subjects) and ETRI-Activity3D~\citep{jang2020etri} (100 subjects) are in Appendix~\ref{app:cross}. \mir{} is trained from scratch on each dataset.

\textbf{Baselines.} Raw skeletons (upper bound on utility and leakage); Gaussian noise ($\sigma{=}0.05$\,m); DMR and PMR~\citep{carr2025privacy}; DisentangledTMR~\citep{carr2026preserving}, described in Section~\ref{sec:related}; and the per-frame residual anonymizer of Moon et al.~\citep{Moon2023SkeletonAnon}, which we train on NTU~60 X-View under our protocol with three seeds.
We use the anonymizer network and objective from the authors' official code (vendored verbatim) with its training loop ported to our data pipeline; the Shift-GCN classifiers of the original are replaced by our raw-trained SGN models (the action model frozen, the identity model trained adversarially from its raw-trained weights), with the published NTU-60 recipe otherwise (SGD, learning rate 0.01, 140 epochs; the reconstruction weight is halved, 10 to 5, because the original averages its MSE over two person slots and our input has one). The vendored code and every deviation will be documented in our code release.
For the retargeting methods, which are our own earlier work, we use our checkpoints from those papers. Every baseline's output is evaluated with the same evaluators as \mir{}. For DisentangledTMR we use all five of its training seeds, so that the comparison is between seed distributions rather than against a single favorable seed.

\textbf{Generation.} Every \mir{} evaluation set is generated through the deployed online path, frame by frame with the streaming state of Section~\ref{sec:method-stream}, not through the batched training path. Section~\ref{sec:res-stream} checks that the two paths agree.

\textbf{Passive adversary A1 (frozen evaluators).} Each evaluator is trained on raw training data and applied unchanged to anonymized test data. For utility we use ten action-recognition families: SGN, Skeleton-MixFormer, CTR-GCN, ST-GCN, 2s-AGCN, MS-G3D, HD-GCN, InfoGCN, FreqMixFormer and SkateFormer. For identity we train the same ten architectures as subject classifiers, plus the gait recognizers GaitGraph2~\citep{teepe2022gaitgraph2} and GaitMixer~\citep{gaitMixer}, adapted from 2D to 25-joint 3D input and trained as closed-set classifiers. SGN and MixFormer are the two primary evaluators throughout.
\emph{1-NN matching.} The attacker embeds a labeled raw gallery (the raw test-split clips) with the raw-trained GaitGraph2 encoder, L2-normalizes the embeddings, and assigns each anonymized test clip the subject of its nearest gallery clip by cosine similarity. We report subject accuracy with matches to the clip's own raw version excluded, which tests whether identity survives as embedding proximity even when a classifier head misses it, and record-linkage accuracy with those matches allowed, which models an attacker who also holds the raw recording.
\emph{Linkage.} Our pairwise same-person verifier~\citep{carr2023linkage}, trained on raw NTU~60 (about \LinkRawAcc\% pair accuracy on raw data), scores balanced same/different-subject pairs (chance \ChanceLink\%) in two settings: both clips anonymized (\emph{within}), and an anonymized clip against a raw reference clip (\emph{cross}).

\textbf{Adaptive adversary A2 (retrained evaluators).} The attacker anonymizes the labeled training split with the released model and trains identity models on it from scratch: SGN and MixFormer for 60 epochs with their released hyperparameters (\emph{retrained}), the same two for 120 epochs with identity-specific tuning (\emph{adaptive}), and GaitGraph2 and GaitMixer with their training recipes. The same retrained SGN and MixFormer pipelines, trained for actions, give retrained AR.

\textbf{Per-subject risk.} Mean RI hides how risk is distributed. For every RI evaluator we compute each subject's top-1 accuracy over that subject's test clips and report the median subject, the worst (most identifiable) subject, the worst-to-mean ratio, and whether the same subjects are most exposed across evaluators.

\textbf{Statistics.} We train the main configuration ($\alpha{=}0.5$, $W{=}\infty$) with five generator seeds and evaluate each seed separately. For each frozen metric we compare \mir{}'s five values with DisentangledTMR's five with Welch's $t$-test (two-sided, unequal variances) and report the difference of means with its 95\% Welch confidence interval, the $p$-value, and Hedges' $g$. With five runs per arm the tests have low power, so we report confidence intervals and effect sizes alongside $p$. The four $p$-values are uncorrected; all four stay below 0.05 under a Bonferroni correction for four comparisons. Retrained-evaluator variance on a fixed generator (three SGN retrainings on the same data) is $\pm$\RetEvalSd\,pp. Sweeps over $\alpha$, $\beta$, window size and ablations use one generator seed (42) and are read as trends, not as significant differences.

\textbf{Streaming checks.} \emph{Equivalence:} we anonymize all \EqNumSeq{} NTU~60 training and test sequences through both the batched and the frame-by-frame path, report the maximum and mean absolute coordinate difference, and evaluate frozen SGN on both outputs. \emph{Long stream:} we feed one continuous stream of \LongT{} synthetic frames through a $W{=}\LongW$ checkpoint and a separate \LongInfT{}-frame stream through $W{=}\infty$, both longer than any training clip, recording cache size, per-frame latency and peak memory. Long-stream runs use a \LongBenchGPU{} GPU; the 64-frame benchmark uses a \BenchGPU. Latency is measured at batch size 1, synchronizing after every frame.

\textbf{Training.} AdamW with learning rates $10^{-4}$ (generator) and $5{\times}10^{-4}$ (heads, discriminator, probe), cosine decay to $10^{-6}$, gradient clipping at 1.0, batch size 64, 30 epochs, final checkpoint; dropout 0.1. Full details are in Appendix~\ref{app:impl}.

%% file: sec_results.tex
\section{Results}
\label{sec:results}

\begin{table*}[t]
\centering
\small
\caption{Passive adversary A1 and drop-in utility on NTU~60 (X-View): evaluators trained on raw data, applied unchanged to anonymized output. AR: action recognition, RI: closed-set re-identification over \NumSubjects{} subjects (chance \ChanceRI\%), both top-1 \%. DisentangledTMR, Moon et al.\ and \mir{} ($\alpha{=}0.5$) are mean{\scriptsize$\,\pm\,$}sd over training seeds (five, three and five); other rows are single runs. $^\dagger$Needs a target skeleton at inference and processes whole clips. $^\S$Per-frame residual, no temporal context; trained by us under our protocol (three seeds). Motion quality in metres; MPJPE: mean per-joint position error to the input; Bone, Smooth, Vel: bone-length, acceleration and velocity errors; motion quality for \mir{} and Moon et al.\ is from seed 42. No row is claimed to dominate another; Table~\ref{tab:sig} tests the two seed distributions.}
\label{tab:frontier}
\setlength{\tabcolsep}{5pt}
\begin{tabular}{l cc cc cccc}
\toprule
& \multicolumn{2}{c}{Frozen SGN} & \multicolumn{2}{c}{Frozen MixFormer} & \multicolumn{4}{c}{Motion quality} \\
\cmidrule(lr){2-3}\cmidrule(lr){4-5}\cmidrule(lr){6-9}
Method & AR\,$\uparrow$ & RI\,$\downarrow$ & AR\,$\uparrow$ & RI\,$\downarrow$ & MPJPE & Bone & Smooth & Vel \\
\midrule
Raw skeletons & \RawSgnAR & \RawSgnRI & \RawMixAR & \RawMixRI & -- & -- & -- & -- \\
Gaussian noise ($\sigma{=}0.05$) & \NoiseSgnAR & \NoiseSgnRI & \NoiseMixAR & \NoiseMixRI & \NoiseMPJPE & \NoiseBone & \NoiseSmooth & \NoiseVel \\
DMR$^\dagger$~\citep{carr2025privacy} & \DmrSgnAR & \DmrSgnRI & \DmrMixAR & \DmrMixRI & \DmrMPJPE & \DmrBone & \DmrSmooth & \DmrVel \\
PMR$^\dagger$~\citep{carr2025privacy} & \PmrSgnAR & \PmrSgnRI & \PmrMixAR & \PmrMixRI & \PmrMPJPE & \PmrBone & \PmrSmooth & \PmrVel \\
DisentangledTMR$^\dagger$~\citep{carr2026preserving}, 5 seeds & \pmsd{\TmrSgnAR}{\TmrSgnARsd} & \pmsd{\TmrSgnRI}{\TmrSgnRIsd} & \pmsd{\TmrMixAR}{\TmrMixARsd} & \pmsd{\TmrMixRI}{\TmrMixRIsd} & \TmrMPJPE & \TmrBone & \TmrSmooth & \TmrVel \\
\quad its best seed & \TmrBestSgnAR & \TmrBestSgnRI & \TmrBestMixAR & \TmrBestMixRI & & & & \\
Moon et al.$^\S$~\citep{Moon2023SkeletonAnon} & \pmsd{\MoonSgnAR}{\MoonSgnARsd} & \pmsd{\MoonSgnRI}{\MoonSgnRIsd} & \pmsd{\MoonMixAR}{\MoonMixARsd} & \pmsd{\MoonMixRI}{\MoonMixRIsd} & \MoonMPJPE & \MoonBone & \MoonSmooth & \MoonVel \\
\midrule
\mir{} $\alpha{=}0.5$, \NumGenSeeds{} seeds & \pmsd{\MainSgnAR}{\MainSgnARsd} & \pmsd{\MainSgnRI}{\MainSgnRIsd} & \pmsd{\MainMixAR}{\MainMixARsd} & \pmsd{\MainMixRI}{\MainMixRIsd} & \MainMPJPE & \MainBone & \MainSmooth & \MainVel \\
\mir{} $\alpha{=}1.0$, seed 42 & \AoneSgnAR & \AoneSgnRI & \AoneMixAR & \AoneMixRI & \AoneMPJPE & \AoneBone & \AoneSmooth & \AoneVel \\
\mir{}-$\beta$ ($\alpha{=}1.0$, $\beta{=}0.9$), seed 42 & \BetaSgnAR & \BetaSgnRI & \BetaMixAR & \BetaMixRI & \BetaMPJPE & \BetaBone & \BetaSmooth & \BetaVel \\
\bottomrule
\end{tabular}
\end{table*}

\subsection{Frozen Frontier Against the Baselines}
\label{sec:res-frontier}

Table~\ref{tab:frontier} reports the passive adversary's identity accuracy and the drop-in utility of each method. The convolutional retargeting methods fail requirement R3: raw-trained SGN recognizes \DmrSgnAR\% (DMR) and \PmrSgnAR\% (PMR) of actions on their output, near the \ChanceAR\% chance level, because their output no longer looks like raw NTU skeletons. Gaussian noise strong enough to lower RI also destroys frozen AR. DisentangledTMR keeps frozen AR high, and its five training seeds span a wide range: \pmsd{\TmrSgnAR}{\TmrSgnARsd}\% AR and \pmsd{\TmrSgnRI}{\TmrSgnRIsd}\% RI on frozen SGN, against \TmrBestSgnAR\% and \TmrBestSgnRI\% for its best seed. Moon et al.'s per-frame residual anonymizer, trained under our protocol, gives \MoonSgnAR\% frozen SGN AR at \MoonSgnRI\% RI: raw-trained classifiers keep working on its output, and so do raw-trained identity models.

\begin{table}[t]
\centering
\small
\caption{Welch tests, \mir{} ($\alpha{=}0.5$, \NumGenSeeds{} generator seeds) minus DisentangledTMR (five training seeds) on the frozen metrics of Table~\ref{tab:frontier}. $\Delta$: difference of means (pp); CI: 95\% Welch interval; $g$: Hedges' $g$. Negative $\Delta$ is better for RI. Uncorrected for four comparisons.}
\label{tab:sig}
\setlength{\tabcolsep}{4pt}
\begin{tabular}{l rcrr}
\toprule
Metric & $\Delta$ & 95\% CI & $p$ & $g$ \\
\midrule
Frozen SGN AR & \SigSgnARdiff & [\SigSgnARlo, \SigSgnARhi] & \SigSgnARp & \SigSgnARg \\
Frozen SGN RI & \SigSgnRIdiff & [\SigSgnRIlo, \SigSgnRIhi] & \SigSgnRIp & \SigSgnRIg \\
Frozen MixFormer AR & \SigMixARdiff & [\SigMixARlo, \SigMixARhi] & \SigMixARp & \SigMixARg \\
Frozen MixFormer RI & \SigMixRIdiff & [\SigMixRIlo, \SigMixRIhi] & \SigMixRIp & \SigMixRIg \\
\bottomrule
\end{tabular}
\end{table}

\mir{}'s main configuration reaches \pmsd{\MainSgnAR}{\MainSgnARsd}\% frozen SGN AR at \pmsd{\MainSgnRI}{\MainSgnRIsd}\% RI. Table~\ref{tab:sig} compares its five seeds with DisentangledTMR's five. All four differences are significant (largest $p=\SigMixRIp$): \mir{} lowers frozen RI (SGN $\Delta=\SigSgnRIdiff$\,pp) and also lowers frozen AR, by much more (SGN $\Delta=\SigSgnARdiff$\,pp). The same pattern holds across ten action-recognition families (Section~\ref{sec:res-suite}). Retrained classifiers recover \RetSgnAR\% AR on the same output (Table~\ref{tab:retrained}), so the low frozen AR reflects a shift away from the distribution raw-trained classifiers expect, not a loss of action information. At $\beta{=}1$, \mir{} therefore does not support drop-in use with existing frozen action models: requirement R3 is not met. Frozen AR also varies across \mir{} seeds (\MainSgnARmin--\MainSgnARmax\%).

\begin{figure}[t]
\centering
\includegraphics[width=\columnwidth]{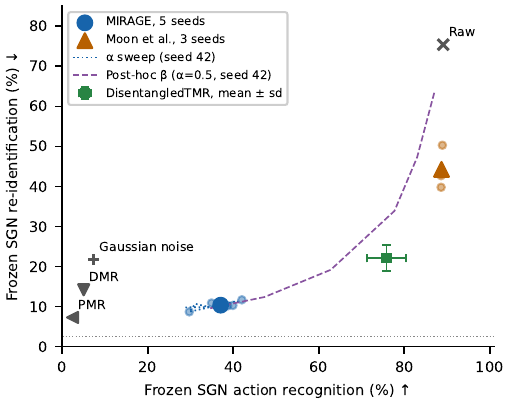}
\caption{Frozen SGN privacy--utility plane on NTU~60. Lower right is better. Individual training seeds and means are shown for \mir{} and Moon et al.; DisentangledTMR is shown as mean$\pm$sd. Retargeting methods without disentanglement sit at near-chance AR. Dashed lines: $\alpha$ sweep and post-hoc $\beta$ for seed 42.}
\Description{Scatter plot of frozen action-recognition accuracy against re-identification accuracy. MIRAGE has lower frozen utility and leakage than DisentangledTMR; Moon et al. retains more of both.}
\label{fig:frontier}
\end{figure}

The $\alpha{=}1.0$ checkpoint (\AoneSgnAR\% AR, \AoneSgnRI\% RI) and its $\beta{=}0.9$ attenuation (\BetaSgnAR\%/\BetaSgnRI\%) are single-seed points inside the main configuration's seed range of frozen AR, so we do not treat them as distinct operating points. Figure~\ref{fig:frontier} places all methods in the AR--RI plane. On motion quality, \mir{}'s mean per-joint deviation from the input (\MainMPJPE\,m) is close to DisentangledTMR's (\TmrMPJPE\,m), whose output has lower bone-length error (\TmrBone{} vs.\ \MainBone) because it adopts a complete target skeleton; Moon et al.'s output stays closer to the input (\MoonMPJPE\,m).

\subsection{Adaptive Adversary}
\label{sec:res-adaptive}

\begin{table}[t]
\centering
\small
\caption{Adaptive adversary A2 and retrained utility: SGN and MixFormer trained from scratch on each method's anonymized training data (60 epochs). \mir{} SGN cells: mean{\scriptsize$\,\pm\,$}sd over generator seeds. $^\dagger$Needs a target skeleton. $^\S$Per-frame residual, trained by us.}
\label{tab:retrained}
\setlength{\tabcolsep}{4pt}
\begin{tabular}{l cc cc}
\toprule
& \multicolumn{2}{c}{Retrained SGN} & \multicolumn{2}{c}{Retrained MixFormer} \\
\cmidrule(lr){2-3}\cmidrule(lr){4-5}
Method & AR\,$\uparrow$ & RI\,$\downarrow$ & AR\,$\uparrow$ & RI\,$\downarrow$ \\
\midrule
Raw & \RawSgnAR & \RawSgnRI & \RawMixAR & \RawMixRI \\
DMR$^\dagger$ & \DmrRetSgnAR & \DmrRetSgnRI & \DmrRetMixAR & \DmrRetMixRI \\
PMR$^\dagger$ & \PmrRetSgnAR & \PmrRetSgnRI & \PmrRetMixAR & \PmrRetMixRI \\
DisentangledTMR$^\dagger$ & \TmrRetSgnAR & \TmrRetSgnRI & \TmrRetMixAR & \TmrRetMixRI \\
Moon et al.$^\S$ & \pmsd{\MoonRetSgnAR}{\MoonRetSgnARsd} & \pmsd{\MoonRetSgnRI}{\MoonRetSgnRIsd} & \MoonRetMixAR & \MoonRetMixRI \\
\mir{} $\alpha{=}0.5$ & \pmsd{\RetSgnAR}{\RetSgnARsd} & \pmsd{\RetSgnRI}{\RetSgnRIsd} & \RetMixAR & \RetMixRI \\
\bottomrule
\end{tabular}
\end{table}

Table~\ref{tab:retrained} gives the main result. A consumer who retrains SGN on \mir{}'s output recognizes \pmsd{\RetSgnAR}{\RetSgnARsd}\% of actions (raw \RawSgnAR\%), so most action utility is recoverable by retraining. An attacker who retrains SGN for identity reaches \pmsd{\RetSgnRI}{\RetSgnRIsd}\% RI on \mir{} output, against \pmsd{\MoonRetSgnRI}{\MoonRetSgnRIsd}\% on the output of Moon et al.'s per-frame residual anonymizer, whose retrained AR is \pmsd{\MoonRetSgnAR}{\MoonRetSgnARsd}\%, and \RawSgnRI\% on raw data. MixFormer gives the same picture (\RetMixAR\%/\RetMixRI\% AR/RI for \mir{}, \MoonRetMixAR\%/\MoonRetMixRI\% for Moon et al.; single seed).

Leakage under A2 nonetheless stays far above chance. Longer, identity-tuned training gives \AdaptSgnRI\% (SGN) and \AdaptMixRI\% (MixFormer). The identity-specialized GaitGraph2, trained on anonymized data, reaches \GaitGraphRetRI\% against its raw ceiling of \GaitGraphRawRI\%, and GaitMixer \GaitMixerRetRI\% against \GaitMixerRawRI\%. The strongest of the \NumAtwoAttackers{} A2 attackers, \WorstRetAttacker{}, reaches \WorstRetRI\% RI (raw \WorstRetRawRI\%, chance \ChanceRI\%). This is \mir{}'s main privacy limitation, and it is why we describe its effect as identity-leakage reduction rather than anonymization.

The retargeting methods leak less under A2 (DisentangledTMR: \TmrRetSgnRI\% RI) because replacing the body with a target's removes the strongest static cue, but a consumer who retrains on their output also recovers far fewer actions (\TmrRetSgnAR\% vs.\ \RetSgnAR\% for \mir{}), and they fail R1 and R2.

\begin{table*}[t]
\centering
\small
\caption{Attacker suite on \mir{} output ($\alpha{=}0.5$; generator seeds: \SuiteSeeds). Raw: evaluator trained and tested on raw data; the eight additional families were re-evaluated together with RI. SGN and MixFormer RI use all 18,932 X-View test clips, including two-person actions; all AR values and the other eight families' RI use the 15,474 single-person-action test clips. A1: raw-trained evaluator on anonymized data. Median and worst subject: per-subject top-1 RI under A1 over the \NumSubjects{} subjects. A2: identity model trained on anonymized data (retrained / adaptive for SGN and MixFormer). Chance: RI \ChanceRI\%, pair accuracy \ChanceLink\%. --: not applicable or not run.}
\label{tab:suite}
\setlength{\tabcolsep}{5pt}
\begin{tabular}{ll cc cc cc c}
\toprule
& & \multicolumn{2}{c}{Raw} & \multicolumn{2}{c}{A1 (frozen)} & \multicolumn{2}{c}{A1 per subject, RI} & A2 \\
\cmidrule(lr){3-4}\cmidrule(lr){5-6}\cmidrule(lr){7-8}\cmidrule(lr){9-9}
Evaluator & Family & AR & RI & AR & RI & median & worst & RI \\
\midrule
SGN~\citep{zhang2020semantics} & semantic GCN & \SuiteSgnRawAR & \SuiteSgnRawRI & \SuiteSgnAR & \SuiteSgnRI & \TailMedSgn & \TailWorstSgn & \AblFullRetRI\,/\,\AdaptSgnRI \\
Skeleton-MixFormer~\citep{xin2023skeleton} & Transformer & \SuiteMixRawAR & \SuiteMixRawRI & \SuiteMixAR & \SuiteMixRI & \TailMedMix & \TailWorstMix & \RetMixRI\,/\,\AdaptMixRI \\
CTR-GCN~\citep{chen2021ctrgcn} & GCN & \SuiteCtrRawAR & \SuiteCtrRawRI & \SuiteCtrAR & \SuiteCtrRI & \TailMedCtr & \TailWorstCtr & -- \\
ST-GCN~\citep{YanXL18} & GCN & \SuiteStgRawAR & \SuiteStgRawRI & \SuiteStgAR & \SuiteStgRI & \TailMedStg & \TailWorstStg & -- \\
2s-AGCN~\citep{shi2019twoStream} & GCN & \SuiteAgcnRawAR & \SuiteAgcnRawRI & \SuiteAgcnAR & \SuiteAgcnRI & \TailMedAgcn & \TailWorstAgcn & -- \\
MS-G3D~\citep{liu2020disentangling} & GCN & \SuiteMsgRawAR & \SuiteMsgRawRI & \SuiteMsgAR & \SuiteMsgRI & \TailMedMsg & \TailWorstMsg & -- \\
HD-GCN~\citep{lee2023hdgcn} & GCN & \SuiteHdRawAR & \SuiteHdRawRI & \SuiteHdAR & \SuiteHdRI & \TailMedHd & \TailWorstHd & -- \\
InfoGCN~\citep{chi2022infogcn} & GCN & \SuiteInfoRawAR & \SuiteInfoRawRI & \SuiteInfoAR & \SuiteInfoRI & \TailMedInfo & \TailWorstInfo & -- \\
FreqMixFormer~\citep{wu2024frequency} & Transformer & \SuiteFreqRawAR & \SuiteFreqRawRI & \SuiteFreqAR & \SuiteFreqRI & \TailMedFreq & \TailWorstFreq & -- \\
SkateFormer~\citep{do2024skateformer} & Transformer & \SuiteSkateRawAR & \SuiteSkateRawRI & \SuiteSkateAR & \SuiteSkateRI & \TailMedSkate & \TailWorstSkate & -- \\
\midrule
GaitGraph2~\citep{teepe2022gaitgraph2} & gait (identity) & -- & \GaitGraphRawRI & -- & \SuiteGaitGraphRI & \TailMedGaitGraph & \TailWorstGaitGraph & \GaitGraphRetRI \\
GaitMixer~\citep{gaitMixer} & gait (identity) & -- & \GaitMixerRawRI & -- & \SuiteGaitMixerRI & \TailMedGaitMixer & \TailWorstGaitMixer & \GaitMixerRetRI \\
\midrule
Mean of the ten action families & & & & \SuiteMeanAR & \SuiteMeanRI & & & \\
\midrule
\multicolumn{2}{l}{1-NN, GaitGraph2 embedding: subject / record linkage} & & & \multicolumn{2}{c}{\KnnTopOne\,/\,\KnnRecordLink} & & & -- \\
\multicolumn{2}{l}{Linkage verifier~\citep{carr2023linkage}: pair acc.\ (AUC), within} & & \LinkRawAcc & \multicolumn{2}{c}{\LinkWithinAcc{} (\LinkWithinAUC)} & & & -- \\
\multicolumn{2}{l}{Linkage verifier~\citep{carr2023linkage}: pair acc.\ (AUC), cross to raw} & & & \multicolumn{2}{c}{\LinkCrossAcc{} (\LinkCrossAUC)} & & & -- \\
\bottomrule
\end{tabular}
\end{table*}

\subsection{Attacker Suite and Per-Subject Risk}
\label{sec:res-suite}

Table~\ref{tab:suite} widens the passive adversary. Across the ten action-recognition families, frozen RI on \mir{} output ranges from \SuiteMinRI\% to \SuiteMaxRI\% (mean \SuiteMeanRI\%) while frozen AR averages \SuiteMeanAR\%, so both the low frozen RI and the low frozen AR are not specific to the two primary evaluators. The identity-specialized gait models are the natural stronger passive attackers: GaitGraph2, whose raw-data ceiling (\GaitGraphRawRI\%) is on par with SGN's, identifies \SuiteGaitGraphRI\% of anonymized clips, and GaitMixer \SuiteGaitMixerRI\%. Matching in GaitGraph2's embedding space instead of using its classifier head gives \KnnTopOne\% subject accuracy, and \KnnRecordLink\% when the attacker also holds each clip's raw recording. The published linkage verifier, \LinkRawAcc\% accurate on raw pairs, reaches \LinkWithinAcc\% pair accuracy when both clips are anonymized and \LinkCrossAcc\% when linking an anonymized clip to a raw reference (chance \ChanceLink\%).

Mean RI does not describe individual risk. Under A1, the most identifiable subject is \TailRatioMin--\TailRatioMax{} times as exposed as the mean, reaching \TailWorstMax\% under \TailWorstMaxEval, while the median subject's RI lies between \TailMedMin\% and \TailMedMax\% across evaluators. For \TailSameTop{} of the twelve evaluators the two most exposed subjects are the same two people, so exposure partly follows the person rather than evaluator noise. A deployer must therefore read mean RI as an average over unequal individual risks: \mir{} lowers but does not equalize them.

\subsection{Online Operation}
\label{sec:res-stream}

\begin{table}[t]
\centering
\small
\caption{Online operation of \mir{} (main configuration, seed 42). (a) Batched vs.\ frame-by-frame output on the \EqNumSeq{} generated sequences. (b) Per-frame cost on a 64-frame clip, $W{=}\infty$, batch size 1, \BenchGPU. (c) Separate long streams on \LongBenchGPU; lengths stated in the table.}
\label{tab:stream}
\setlength{\tabcolsep}{3pt}
\begin{tabular}{l cc}
\toprule
\multicolumn{3}{l}{\emph{(a) Batch--stream equivalence}} \\
Max / mean abs.\ coordinate difference (m) & \multicolumn{2}{c}{\EqMaxAbs{} / \EqMeanAbs} \\
 & Batched & Streamed \\
Frozen SGN AR / RI (\%) & \EqBatchSgnAR{} / \EqBatchSgnRI & \EqStreamSgnAR{} / \EqStreamSgnRI \\
\midrule
\multicolumn{3}{l}{\emph{(b) Per-frame cost, $T{=}64$}} \\
 & Batched & Streamed \\
ms per frame (p95) & \BatchMsFrame & \StreamMsFrame{} (\StreamPninetyfive) \\
Frames per second & \BatchFPS & \StreamFPS \\
Peak GPU memory (MB) & \BatchPeakMB & \StreamPeakMB \\
\midrule
\multicolumn{3}{l}{\emph{(c) Long streams, \LongBenchGPU}} \\
 & $W{=}\LongW$ & $W{=}\infty$ \\
Frames & \LongT & \LongInfT \\
Cache entries per attention at end & \LongCacheEntries & grows with $t$ \\
Cache size at end (KiB) & \LongCacheKB & -- \\
ms per frame, first / last 1k frames & \LongMsStart{} / \LongMsEnd & -- / \LongInfMsEnd \\
Peak memory, start / end (MB) & \LongPeakMBStart{} / \LongPeakMBEnd & -- / \LongInfPeakMBEnd \\
\bottomrule
\end{tabular}
\end{table}

\textbf{Equivalence.} Table~\ref{tab:stream}(a) checks the property of Section~\ref{sec:method-causal} on real data. Over all \EqNumSeq{} training and test sequences, batched and frame-by-frame outputs differ by at most \EqMaxAbs\,m (mean \EqMeanAbs\,m), and frozen SGN gives \EqStreamSgnAR\%/\EqStreamSgnRI\% AR/RI on streamed output against \EqBatchSgnAR\%/\EqBatchSgnRI\% on batched output. All \mir{} accuracy and motion-quality measurements in this paper use streamed output, so they describe the function that runs online.

\textbf{Bounded memory.} On a stream of \LongT{} frames with $W{=}\LongW$, every attention cache holds \LongCacheEntries{} entries at the end, the total cache is \LongCacheKB\,KiB, and per-frame latency (\LongMsStart{}\,ms over the first thousand frames, \LongMsEnd{}\,ms over the last) and peak memory (\LongPeakMBStart{} vs.\ \LongPeakMBEnd\,MB) do not grow. With $W{=}\infty$, a separate \LongInfT{}-frame stream grows to \LongInfMsEnd\,ms per frame and \LongInfPeakMBEnd\,MB at the end, as Section~\ref{sec:method-stream} predicts. Bounded state therefore requires a finite window.

\textbf{Choosing a window.} Table~\ref{tab:window} retrains \mir{} with finite windows (seed 42). A window $W$ bounds the state and limits the receptive field to $R=(L_E+L_D)(W-1)+1$ frames: 25, 57, 121, 249 and 505 frames for $W=4,8,16,32,64$. The operating point moves with $W$: retrained RI is \WinFourRetRI\%, \WinEightRetRI\% and \WinSixteenRetRI\% for $W=4,8,16$ against \WinInfRetRI\% for $W{=}\infty$, at similar retrained AR, so a deployer who needs bounded memory should choose $W$ from this table rather than assume that the $W{=}\infty$ results carry over.

\begin{table}[t]
\centering
\small
\caption{Context window $W$ on NTU~60 ($\alpha{=}0.5$, seed 42). SGN evaluators.}
\label{tab:window}
\begin{tabular}{r cc cc}
\toprule
& \multicolumn{2}{c}{Frozen (A1)} & \multicolumn{2}{c}{Retrained (A2)} \\
\cmidrule(lr){2-3}\cmidrule(lr){4-5}
$W$ & AR & RI & AR & RI \\
\midrule
4 & \WinFourFrzAR & \WinFourFrzRI & \WinFourRetAR & \WinFourRetRI \\
8 & \WinEightFrzAR & \WinEightFrzRI & \WinEightRetAR & \WinEightRetRI \\
16 & \WinSixteenFrzAR & \WinSixteenFrzRI & \WinSixteenRetAR & \WinSixteenRetRI \\
32 & \WinThirtyTwoFrzAR & \WinThirtyTwoFrzRI & \WinThirtyTwoRetAR & \WinThirtyTwoRetRI \\
64 & \WinSixtyFourFrzAR & \WinSixtyFourFrzRI & \WinSixtyFourRetAR & \WinSixtyFourRetRI \\
$\infty$ & \WinInfFrzAR & \WinInfFrzRI & \WinInfRetAR & \WinInfRetRI \\
\bottomrule
\end{tabular}
\end{table}

\textbf{Latency.} On a 64-frame clip, frame-by-frame inference takes \StreamMsFrame\,ms per frame (\StreamFPS{} frames per second), within the 33\,ms budget of 30\,fps capture for a 64-frame clip; with $W{=}\infty$ per-frame latency grows with stream length (Table~\ref{tab:stream}c). We present this as a property of our implementation on one GPU, not as a contribution: it depends on hardware, batch size and serving choices.

\begin{table}[t]
\centering
\small
\caption{Size and cost. Params: deployed generator (M). ms/clip: one 63-frame clip, batch size 1. Baseline timings are historical V100S measurements; MIRAGE timings use \CostBenchGPU. GPU-h are per-run training costs on the original hardware; these are not controlled speed comparisons. Speed depends on serving choices (caching, batching) that are not specific to any method; only the last two columns are structural.}
\label{tab:cost}
\setlength{\tabcolsep}{2.5pt}
\begin{tabular}{l rrr cc}
\toprule
Method & Params & GPU-h & ms/clip & Target & Online \\
\midrule
DMR & \CostDmrParams & \CostDmrTrainH & \CostDmrMsSeq & yes & no \\
PMR & \CostPmrParams & \CostPmrTrainH & \CostPmrMsSeq & yes & no \\
DisentangledTMR & \CostTmrParams & \CostTmrTrainH & \CostTmrMsSeq & yes & no \\
\mir{}, batched & \CostMirageParams & \CostMirageTrainH & \CostMirageBatchMsSeq & no & no \\
\mir{}, streamed & \CostMirageParams & & \CostMirageStreamMsSeq & no & yes \\
\quad without caches & \CostMirageParams & & \CostMirageNoCacheMsSeq & no & yes \\
\bottomrule
\end{tabular}
\end{table}

\textbf{Size and speed.} Table~\ref{tab:cost} separates structural differences from serving choices. \mir{}'s generator has \CostMirageParams{}M parameters, about \CostParamRatio{} times fewer than DisentangledTMR's \CostTmrParams{}M. DisentangledTMR's \CostTmrMsSeq\,ms per clip comes from a 63-step autoregressive decode without key/value caching; caching is a standard optimization that it could adopt. The two streamed timings are similar and were measured on different GPUs, so we draw no speed conclusion. The rows for \mir{} with and without caches, measured on the same \CostBenchGPU, isolate its own caching benefit. Its batched 64-frame forward pass uses \CostMirageGflopsSeq{} GFLOPs. The structural difference is the last two columns: DisentangledTMR cannot emit a frame before it has the whole clip and a target skeleton, under any serving scheme.

\subsection{Operating Points and Components}
\label{sec:res-ops}

\begin{table}[t]
\centering
\small
\caption{Privacy weight $\alpha$ (top) and component ablation (bottom), seed 42; each ablation row removes one component from the full model. Frozen and retrained evaluator: SGN. Single runs: read trends, not small differences.}
\label{tab:ablation}
\label{tab:alpha}
\setlength{\tabcolsep}{4pt}
\begin{tabular}{l cc cc}
\toprule
& \multicolumn{2}{c}{Frozen (A1)} & \multicolumn{2}{c}{Retrained (A2)} \\
\cmidrule(lr){2-3}\cmidrule(lr){4-5}
Configuration & AR\,$\uparrow$ & RI\,$\downarrow$ & AR\,$\uparrow$ & RI\,$\downarrow$ \\
\midrule
$\alpha{=}0.10$ & \AlphaPtOneFrzAR & \AlphaPtOneFrzRI & \AlphaPtOneRetAR & \AlphaPtOneRetRI \\
$\alpha{=}0.25$ & \AlphaPtTwoFiveFrzAR & \AlphaPtTwoFiveFrzRI & \AlphaPtTwoFiveRetAR & \AlphaPtTwoFiveRetRI \\
$\alpha{=}0.50$ (main) & \AlphaPtFiveFrzAR & \AlphaPtFiveFrzRI & \AlphaPtFiveRetAR & \AlphaPtFiveRetRI \\
$\alpha{=}0.75$ & \AlphaPtSevenFiveFrzAR & \AlphaPtSevenFiveFrzRI & \AlphaPtSevenFiveRetAR & \AlphaPtSevenFiveRetRI \\
$\alpha{=}1.00$ & \AlphaOneFrzAR & \AlphaOneFrzRI & \AlphaOneRetAR & \AlphaOneRetRI \\
$\alpha{=}2.00$ & \AlphaTwoFrzAR & \AlphaTwoFrzRI & \AlphaTwoRetAR & \AlphaTwoRetRI \\
\midrule
Full model ($\alpha{=}0.5$) & \AblFullFrzAR & \AblFullFrzRI & \AblFullRetAR & \AblFullRetRI \\
No contrastive probe & \AblNoConFrzAR & \AblNoConFrzRI & \AblNoConRetAR & \AblNoConRetRI \\
No distribution discr. & \AblNoDdFrzAR & \AblNoDdFrzRI & \AblNoDdRetAR & \AblNoDdRetRI \\
Neither of the two & \AblNeitherFrzAR & \AblNeitherFrzRI & \AblNeitherRetAR & \AblNeitherRetRI \\
No output ID adversary & \AblNoOutIdFrzAR & \AblNoOutIdFrzRI & \AblNoOutIdRetAR & \AblNoOutIdRetRI \\
No action heads & \AblNoActFrzAR & \AblNoActFrzRI & \AblNoActRetAR & \AblNoActRetRI \\
No latent ID adversary & \AblNoLatIdFrzAR & \AblNoLatIdFrzRI & \AblNoLatIdRetAR & \AblNoLatIdRetRI \\
No temporal loss & \AblNoTempFrzAR & \AblNoTempFrzRI & \AblNoTempRetAR & \AblNoTempRetRI \\
No bone-length loss & \AblNoBoneFrzAR & \AblNoBoneFrzRI & \AblNoBoneRetAR & \AblNoBoneRetRI \\
No adversarial warm-up & \AblNoProgFrzAR & \AblNoProgFrzRI & \AblNoProgRetAR & \AblNoProgRetRI \\
\bottomrule
\end{tabular}
\end{table}

\textbf{Privacy weight.} Raising $\alpha$ from 0.1 to 1.0 lowers retrained RI from \AlphaPtOneRetRI\% to \AlphaOneRetRI\% ($\alpha{=}2.0$: \AlphaTwoRetRI\%), while retrained AR stays between \AlphaRetARmin\% and \AlphaRetARmax\% (Table~\ref{tab:alpha}). Frozen AR is low at every $\alpha$ (\AlphaTwoFrzAR--\AlphaPtOneFrzAR\%), so the privacy weight does not restore drop-in use.

\textbf{Residual scale.} Frozen AR comes back only when the residual shrinks. Scaling the $\alpha{=}0.5$ residual to $\beta{=}0.3$ at inference gives \res{beta/0.5/0.3/FrzAR}\% frozen SGN AR at \res{beta/0.5/0.3/FrzRI}\% frozen RI and \res{beta/0.5/0.3/RetRI}\% retrained RI (seed 42): a post-hoc dial that trades privacy back for frozen utility, not a better operating point (full grid in Appendix~\ref{app:beta}). Fixing $\beta{=}0.3$ during training instead (\BtrainNumSeeds{} seeds) raises frozen AR to \pmsd{\BtrainFrzAR}{\BtrainFrzARsd}\% but also raises retrained RI to \pmsd{\BtrainRetRI}{\BtrainRetRIsd}\% and adaptive RI to \pmsd{\BtrainAdaptRI}{\BtrainAdaptRIsd}\% at the same retrained AR (\pmsd{\BtrainRetAR}{\BtrainRetARsd}\%); under retraining it is dominated by $\beta{=}1$.

\textbf{Components.} Frozen AR and adaptive leakage move together across components. Removing the distribution discriminator lowers frozen AR from \AblFullFrzAR\% to \AblNoDdFrzAR\% (\AblNeitherFrzAR\% without the contrastive probe as well), and raises retrained RI to \AblNoDdRetRI\%. Every other ablation raises frozen AR, most of all without the action heads (\AblNoActFrzAR\%) or without the output identity adversary (\AblNoOutIdFrzAR\%), and these two also leak most under retraining (\AblNoActRetRI\% and \AblNoOutIdRetRI\% retrained RI). The full model has the lowest retrained RI of all ablation rows (\AblFullRetRI\%). The learned residual is not uniform over the body: it is smallest on the spine (\PertSpine\,m mean per-joint displacement) and largest on the legs (Appendix~\ref{app:qual}), although no body-part-specific perturbation targets are supplied.

\textbf{Other datasets.} Trained from scratch on NTU~120 (106 subjects), \mir{} gives \CrossNtuOTRetAR\% retrained AR and \CrossNtuOTRetRI\% retrained RI (raw: \CrossRawNtuOTAR\% and \CrossRawNtuOTRI\%); on ETRI (100 subjects), \CrossEtriRetAR\% and \CrossEtriRetRI\% (raw: \CrossRawEtriAR\% and \CrossRawEtriRI\%). Appendix~\ref{app:cross} lists frozen results and baselines.

%% file: sec_discussion.tex
\section{Discussion}
\label{sec:discussion}

\textbf{What the evidence supports.} \mir{} meets R1, R2 and, with a finite window, R4; at full residual scale it does not meet R3. Under retraining, the setting we treat as the headline, it keeps \RetSgnAR\% action accuracy and holds retrained SGN identification to \RetSgnRI\%, against \MoonRetSgnRI\% for the per-frame residual anonymizer of Moon et al.~\citep{Moon2023SkeletonAnon} at \MoonRetSgnAR\% AR. Against raw-trained models it lowers identification below the strongest retargeting baseline (\MainSgnRI\% vs.\ \TmrSgnRI\% frozen SGN RI), and the low RI holds across ten architecture families and two gait models, but raw-trained action classifiers fail on the same output (\MainSgnAR\% frozen AR). The contribution is causal temporal de-identification with lower adaptive leakage than a per-frame residual, measured against adaptive attackers; it is not a drop-in replacement for raw input to existing frozen models. The comparison with Moon et al.\ does not isolate temporal context from \mir{}'s larger perturbation and different objective (Table~\ref{tab:frontier}).

\textbf{When to use it.} \mir{} fits a curator who releases a live stream to consumers that can retrain their action models on released data, and who wants lower leakage to output-trained adversaries than a per-frame residual gives, accepting that the strongest such adversary still identifies \WorstRetRI\% of clips. Consumers who must keep using raw-trained classifiers unchanged are not served at $\beta{=}1$; a smaller post-hoc $\beta$ restores frozen accuracy at the cost of higher identification. If the anonymizer's weights or query access are available to the adversary, \mir{} alone is not sufficient protection; it should be combined with access control on the anonymizer, contractual limits on consumers, or retention limits, and its residual risk should be measured with retrained attackers. If neither target-freeness nor causality is required, retargeting gives lower A2 leakage at a large cost in retrained utility.

\textbf{Why adaptive leakage remains.} The residual keeps the output in the input's coordinate space and changes it by a consistent, learnable transformation. Identity cues that are hidden from raw-trained identity models but survive that transformation are exactly what an attacker who trains on the output can exploit. The ablations show the tension directly: the components whose removal raises frozen action accuracy also raise retrained identification. Lowering adaptive leakage while restoring frozen utility is the main open problem, and adaptive-first evaluation~\citep{hanisch2024false} is the right yardstick for it.

\textbf{Limitations.}
(i) \emph{Empirical only.} All privacy statements are measurements against the attackers we ran; there is no formal guarantee, and stronger attackers, such as white-box inversion of the released weights or ensembles across attack families, may recover more.
(ii) \emph{Adaptive leakage and drop-in use.} Under A2, RI reaches \WorstRetRI\% against \WorstRetRawRI\% on raw data; the method should not be described to data subjects as anonymization. At $\beta{=}1$ raw-trained action classifiers do not transfer to its output.
(iii) \emph{Data.} All main results use NTU RGB+D~60: laboratory Kinect~v2 recordings, one skeleton per clip, of \NumSubjects{} subjects with limited demographic diversity. Effectiveness may differ for in-the-wild pose estimates, other skeleton formats, larger populations, and under-represented body types and movement patterns. The NTU~120 and ETRI runs use a single seed.
(iv) \emph{Closed set.} RI measures identification among a known gallery of \NumSubjects{} people; open-set identification and verification are evaluated only through the linkage diagnostic.
(v) \emph{Statistics.} Five seeds per arm give low power; the Moon et al.\ baseline has \MoonNumSeeds{} seeds and no significance test, the retrained MixFormer and adaptive attackers use one generator seed, and the sweeps and ablations are single-seed.
(vi) \emph{Coverage of related methods.} We did not evaluate deep motion masking~\citep{nair2024deep} or Pantomime~\citep{hanisch2026pantomime} under our protocol.
(vii) \emph{Streaming scope.} Our main configuration uses $W{=}\infty$, whose state grows with stream length; bounded memory requires a finite window, which changes the operating point (Table~\ref{tab:window}). Latency was measured on a server GPU, not on capture devices.

\section{Conclusion}
\label{sec:conclusion}

Per-frame residual maps already made skeleton de-identification target-free and causal; \mir{} adds causal temporal context under the same constraints. Its residual encoder--decoder with one causal mask for every attention computes the same function offline and frame by frame and keeps state bounded under a finite window. When consumers retrain, it keeps most action utility and leaks less identity than a per-frame residual anonymizer trained under the same protocol. Raw-trained action classifiers, however, do not transfer to its output at full residual scale, and adaptive attackers still identify far more people than chance. The result is a measured reduction of identity leakage for consumers who retrain, and a clear target for future work: lowering adaptive leakage while keeping raw-trained classifiers working.

%% file: appendix.tex
\section{Implementation Details}
\label{app:impl}

\textbf{Architecture.} $L_E{=}L_D{=}4$ pre-norm layers, $d{=}256$, 8 heads of dimension 32, feed-forward width 1024 with GELU, dropout 0.1, rotary embeddings with base $10^4$ applied to queries and keys of every attention. The deployed generator has \CostMirageParams{}M parameters (encoder, decoder, start token). The four supervision heads add 0.35M parameters, the contrastive probe 26,240 ($75{\to}128{\to}128$), and the distribution discriminator is a three-layer temporal CNN with spectral normalization and a two-layer head; all are discarded at deployment.

\textbf{Data handling.} Clips are processed at their original length (up to 300 frames) with padding masks; masked keys are excluded from attention and padded frames from every loss. Trailing padding does not change the output for valid frames, which a unit test pins. Downstream evaluators apply their own temporal sampling.

\textbf{Optimization.} AdamW~\citep{loshchilov2017decoupled}; learning rate $10^{-4}$ for the generator and $5{\times}10^{-4}$ for the heads, discriminator and probe; cosine decay to $10^{-6}$; gradient-norm clipping at 1.0 per parameter group; batch size 64; 30 epochs; final checkpoint (no selection on validation privacy). The adversarial identity term is ramped linearly over the first $E_w{=}10$ epochs. Generator seeds for the main configuration: 42, 123, 7, 100, 200; all sweeps use seed 42.

\textbf{Augmentation.} Random 3D rotation ($\theta{=}0.3$\,rad), left--right mirroring ($p{=}0.5$), Gaussian joint noise ($\sigma{=}0.01$), uniform scaling (0.9--1.1), and random shear ($\sigma{=}0.02$); each training clip is repeated five times per epoch with independent augmentation.

\textbf{Losses.} With $\mathcal V$ the valid frames, $\mathcal V'$ the valid consecutive pairs, $\Delta\mathbf x_t=\mathbf x_t-\mathbf x_{t-1}$, and $\mathcal B$ the 24 bones of the NTU skeleton:
\begin{align*}
\mathcal L_{\mathrm{rec}}&=\tfrac{1}{|\mathcal V|}\textstyle\sum_{t\in\mathcal V}\|\hat{\mathbf s}_t-\mathbf s_t\|^2,\\
\mathcal L_{\mathrm{temp}}&=\tfrac{1}{|\mathcal V'|}\textstyle\sum_{t\in\mathcal V'}\|\Delta\hat{\mathbf s}_t-\Delta\mathbf s_t\|^2,\\
\mathcal L_{\mathrm{bone}}&=\tfrac{1}{T|\mathcal B|}\textstyle\sum_t\sum_{(u,v)\in\mathcal B}\bigl(\|\hat{\mathbf s}^u_t-\hat{\mathbf s}^v_t\|-\|\mathbf s^u_t-\mathbf s^v_t\|\bigr)^2 .
\end{align*}
The discriminator is trained with binary cross-entropy on raw (label 1) versus detached anonymized (label 0) sequences, and the generator term is $\mathcal L^G_{\mathrm{dd}}=-\log\sigma(\mathrm{DD}(\hat{\mathbf S}))$. Identity and action losses are cross-entropies on the heads' logits; the generator negates the identity terms.

\textbf{Verification of the causal implementation.} The tests, which we will release with the code, check, for windowed and unwindowed models, that (a) changing frames after $t$ leaves batched outputs up to $t$ unchanged, (b) frame-by-frame and batched outputs agree to $10^{-4}$, (c) a 400-frame stream (longer than the 300-frame training clips) runs with every cache holding exactly $W$ entries, and (d) trailing padding does not change valid outputs.

\section{MIRAGE-$\beta$: Residual Scaling at Inference}
\label{app:beta}

Because the output is $\mathbf s_t+\beta\boldsymbol\delta_t$, a deployer can attenuate the learned residual without retraining. Table~\ref{tab:beta-posthoc} applies $\beta\in\{0.1,\dots,0.9\}$ at inference to four trained checkpoints; Table~\ref{tab:beta-train} instead fixes $\beta$ during training of the $\alpha{=}0.5$ model. Within each $\alpha$, larger $\beta$ lowers both frozen RI and frozen AR, tracing a frontier; at $\beta{=}1$ (the main configuration) frozen AR is low, and only smaller $\beta$ restores it, at higher RI. Across $\alpha$ the curves overlap. All cells are single runs (seed 42), so neighboring cells should not be ranked against each other or against a baseline's single seed.

\begin{table}[!htbp]
\centering
\footnotesize
\caption{Post-hoc $\alpha\times\beta$ grid on NTU~60. Retr: SGN retrained on anonymized data (A2). Frz: raw-trained SGN and MixFormer (A1); frozen values are recorded to one decimal, so their second decimal is always zero.}
\label{tab:beta-posthoc}
\setlength{\tabcolsep}{3pt}
\begin{tabular}{cc cc cc cc}
\toprule
& & \multicolumn{2}{c}{Retr SGN} & \multicolumn{2}{c}{Frz SGN} & \multicolumn{2}{c}{Frz MixF} \\
\cmidrule(lr){3-4}\cmidrule(lr){5-6}\cmidrule(lr){7-8}
$\alpha$ & $\beta$ & AR & RI & AR & RI & AR & RI \\
\midrule
0.5 & 0.1 & \res{beta/0.5/0.1/RetAR} & \res{beta/0.5/0.1/RetRI} & \res{beta/0.5/0.1/FrzAR} & \res{beta/0.5/0.1/FrzRI} & \res{beta/0.5/0.1/MixAR} & \res{beta/0.5/0.1/MixRI} \\
0.5 & 0.2 & \res{beta/0.5/0.2/RetAR} & \res{beta/0.5/0.2/RetRI} & \res{beta/0.5/0.2/FrzAR} & \res{beta/0.5/0.2/FrzRI} & \res{beta/0.5/0.2/MixAR} & \res{beta/0.5/0.2/MixRI} \\
0.5 & 0.3 & \res{beta/0.5/0.3/RetAR} & \res{beta/0.5/0.3/RetRI} & \res{beta/0.5/0.3/FrzAR} & \res{beta/0.5/0.3/FrzRI} & \res{beta/0.5/0.3/MixAR} & \res{beta/0.5/0.3/MixRI} \\
0.5 & 0.5 & \res{beta/0.5/0.5/RetAR} & \res{beta/0.5/0.5/RetRI} & \res{beta/0.5/0.5/FrzAR} & \res{beta/0.5/0.5/FrzRI} & \res{beta/0.5/0.5/MixAR} & \res{beta/0.5/0.5/MixRI} \\
0.5 & 0.7 & \res{beta/0.5/0.7/RetAR} & \res{beta/0.5/0.7/RetRI} & \res{beta/0.5/0.7/FrzAR} & \res{beta/0.5/0.7/FrzRI} & \res{beta/0.5/0.7/MixAR} & \res{beta/0.5/0.7/MixRI} \\
0.5 & 0.9 & \res{beta/0.5/0.9/RetAR} & \res{beta/0.5/0.9/RetRI} & \res{beta/0.5/0.9/FrzAR} & \res{beta/0.5/0.9/FrzRI} & \res{beta/0.5/0.9/MixAR} & \res{beta/0.5/0.9/MixRI} \\
\midrule
0.75 & 0.1 & \res{beta/0.75/0.1/RetAR} & \res{beta/0.75/0.1/RetRI} & \res{beta/0.75/0.1/FrzAR} & \res{beta/0.75/0.1/FrzRI} & \res{beta/0.75/0.1/MixAR} & \res{beta/0.75/0.1/MixRI} \\
0.75 & 0.2 & \res{beta/0.75/0.2/RetAR} & \res{beta/0.75/0.2/RetRI} & \res{beta/0.75/0.2/FrzAR} & \res{beta/0.75/0.2/FrzRI} & \res{beta/0.75/0.2/MixAR} & \res{beta/0.75/0.2/MixRI} \\
0.75 & 0.3 & \res{beta/0.75/0.3/RetAR} & \res{beta/0.75/0.3/RetRI} & \res{beta/0.75/0.3/FrzAR} & \res{beta/0.75/0.3/FrzRI} & \res{beta/0.75/0.3/MixAR} & \res{beta/0.75/0.3/MixRI} \\
0.75 & 0.5 & \res{beta/0.75/0.5/RetAR} & \res{beta/0.75/0.5/RetRI} & \res{beta/0.75/0.5/FrzAR} & \res{beta/0.75/0.5/FrzRI} & \res{beta/0.75/0.5/MixAR} & \res{beta/0.75/0.5/MixRI} \\
0.75 & 0.7 & \res{beta/0.75/0.7/RetAR} & \res{beta/0.75/0.7/RetRI} & \res{beta/0.75/0.7/FrzAR} & \res{beta/0.75/0.7/FrzRI} & \res{beta/0.75/0.7/MixAR} & \res{beta/0.75/0.7/MixRI} \\
0.75 & 0.9 & \res{beta/0.75/0.9/RetAR} & \res{beta/0.75/0.9/RetRI} & \res{beta/0.75/0.9/FrzAR} & \res{beta/0.75/0.9/FrzRI} & \res{beta/0.75/0.9/MixAR} & \res{beta/0.75/0.9/MixRI} \\
\midrule
1.0 & 0.1 & \res{beta/1.0/0.1/RetAR} & \res{beta/1.0/0.1/RetRI} & \res{beta/1.0/0.1/FrzAR} & \res{beta/1.0/0.1/FrzRI} & \res{beta/1.0/0.1/MixAR} & \res{beta/1.0/0.1/MixRI} \\
1.0 & 0.2 & \res{beta/1.0/0.2/RetAR} & \res{beta/1.0/0.2/RetRI} & \res{beta/1.0/0.2/FrzAR} & \res{beta/1.0/0.2/FrzRI} & \res{beta/1.0/0.2/MixAR} & \res{beta/1.0/0.2/MixRI} \\
1.0 & 0.3 & \res{beta/1.0/0.3/RetAR} & \res{beta/1.0/0.3/RetRI} & \res{beta/1.0/0.3/FrzAR} & \res{beta/1.0/0.3/FrzRI} & \res{beta/1.0/0.3/MixAR} & \res{beta/1.0/0.3/MixRI} \\
1.0 & 0.5 & \res{beta/1.0/0.5/RetAR} & \res{beta/1.0/0.5/RetRI} & \res{beta/1.0/0.5/FrzAR} & \res{beta/1.0/0.5/FrzRI} & \res{beta/1.0/0.5/MixAR} & \res{beta/1.0/0.5/MixRI} \\
1.0 & 0.7 & \res{beta/1.0/0.7/RetAR} & \res{beta/1.0/0.7/RetRI} & \res{beta/1.0/0.7/FrzAR} & \res{beta/1.0/0.7/FrzRI} & \res{beta/1.0/0.7/MixAR} & \res{beta/1.0/0.7/MixRI} \\
1.0 & 0.9 & \res{beta/1.0/0.9/RetAR} & \res{beta/1.0/0.9/RetRI} & \res{beta/1.0/0.9/FrzAR} & \res{beta/1.0/0.9/FrzRI} & \res{beta/1.0/0.9/MixAR} & \res{beta/1.0/0.9/MixRI} \\
\midrule
2.0 & 0.1 & \res{beta/2.0/0.1/RetAR} & \res{beta/2.0/0.1/RetRI} & \res{beta/2.0/0.1/FrzAR} & \res{beta/2.0/0.1/FrzRI} & \res{beta/2.0/0.1/MixAR} & \res{beta/2.0/0.1/MixRI} \\
2.0 & 0.2 & \res{beta/2.0/0.2/RetAR} & \res{beta/2.0/0.2/RetRI} & \res{beta/2.0/0.2/FrzAR} & \res{beta/2.0/0.2/FrzRI} & \res{beta/2.0/0.2/MixAR} & \res{beta/2.0/0.2/MixRI} \\
2.0 & 0.3 & \res{beta/2.0/0.3/RetAR} & \res{beta/2.0/0.3/RetRI} & \res{beta/2.0/0.3/FrzAR} & \res{beta/2.0/0.3/FrzRI} & \res{beta/2.0/0.3/MixAR} & \res{beta/2.0/0.3/MixRI} \\
2.0 & 0.5 & \res{beta/2.0/0.5/RetAR} & \res{beta/2.0/0.5/RetRI} & \res{beta/2.0/0.5/FrzAR} & \res{beta/2.0/0.5/FrzRI} & \res{beta/2.0/0.5/MixAR} & \res{beta/2.0/0.5/MixRI} \\
2.0 & 0.7 & \res{beta/2.0/0.7/RetAR} & \res{beta/2.0/0.7/RetRI} & \res{beta/2.0/0.7/FrzAR} & \res{beta/2.0/0.7/FrzRI} & \res{beta/2.0/0.7/MixAR} & \res{beta/2.0/0.7/MixRI} \\
2.0 & 0.9 & \res{beta/2.0/0.9/RetAR} & \res{beta/2.0/0.9/RetRI} & \res{beta/2.0/0.9/FrzAR} & \res{beta/2.0/0.9/FrzRI} & \res{beta/2.0/0.9/MixAR} & \res{beta/2.0/0.9/MixRI} \\
\bottomrule
\end{tabular}
\end{table}

\begin{table}[!htbp]
\centering
\footnotesize
\caption{Train-time $\beta$ ($\alpha{=}0.5$, $\beta$ fixed during training). Columns as in Table~\ref{tab:beta-posthoc}.}
\label{tab:beta-train}
\setlength{\tabcolsep}{3pt}
\begin{tabular}{c cc cc cc}
\toprule
& \multicolumn{2}{c}{Retr SGN} & \multicolumn{2}{c}{Frz SGN} & \multicolumn{2}{c}{Frz MixF} \\
\cmidrule(lr){2-3}\cmidrule(lr){4-5}\cmidrule(lr){6-7}
$\beta$ & AR & RI & AR & RI & AR & RI \\
\midrule
0.2 & \res{beta/train/0.2/RetAR} & \res{beta/train/0.2/RetRI} & \res{beta/train/0.2/FrzAR} & \res{beta/train/0.2/FrzRI} & \res{beta/train/0.2/MixAR} & \res{beta/train/0.2/MixRI} \\
0.3 & \res{beta/train/0.3/RetAR} & \res{beta/train/0.3/RetRI} & \res{beta/train/0.3/FrzAR} & \res{beta/train/0.3/FrzRI} & \res{beta/train/0.3/MixAR} & \res{beta/train/0.3/MixRI} \\
0.5 & \res{beta/train/0.5/RetAR} & \res{beta/train/0.5/RetRI} & \res{beta/train/0.5/FrzAR} & \res{beta/train/0.5/FrzRI} & \res{beta/train/0.5/MixAR} & \res{beta/train/0.5/MixRI} \\
0.7 & \res{beta/train/0.7/RetAR} & \res{beta/train/0.7/RetRI} & \res{beta/train/0.7/FrzAR} & \res{beta/train/0.7/FrzRI} & \res{beta/train/0.7/MixAR} & \res{beta/train/0.7/MixRI} \\
\bottomrule
\end{tabular}
\end{table}

\section{Cross-Dataset Results}
\label{app:cross}

We train \mir{} from scratch, with the main recipe and seed 42, on NTU RGB+D~120~\citep{liu2019ntu}, which has 106 subjects (RI chance 0.9\%), and on ETRI-Activity3D~\citep{jang2020etri}, which has 100 subjects (RI chance 1\%). The evaluated files contain 120 NTU actions and 51 ETRI actions, rather than the primary NTU-60 single-person action inventory. Table~\ref{tab:cross} uses SGN evaluators. DMR and PMR NTU~120 numbers are those published in our earlier work~\citep{carr2025privacy} with a slightly different evaluator setup, so they are not strictly comparable; neither method has published ETRI results. Raw and DisentangledTMR numbers come from our earlier work~\citep{carr2026preserving}.

\begin{table}[!htbp]
\centering
\small
\caption{Cross-dataset results (\%). Retrained: SGN trained on anonymized data (A2). Frozen: raw-trained classifiers (A1). A dash omits an action score whose frozen classifier has an incompatible label-space size; identity heads have matching cardinalities. DMR and PMR have no published ETRI results (dashes). $^\dagger$Needs a target skeleton. $^\ddagger$Published numbers, different evaluator setup.}
\label{tab:cross}
\setlength{\tabcolsep}{4pt}
\begin{tabular}{l cc cc}
\toprule
& \multicolumn{2}{c}{NTU~120} & \multicolumn{2}{c}{ETRI} \\
\cmidrule(lr){2-3}\cmidrule(lr){4-5}
Method & AR & RI & AR & RI \\
\midrule
Raw & \CrossRawNtuOTAR & \CrossRawNtuOTRI & \CrossRawEtriAR & \CrossRawEtriRI \\
DMR$^{\dagger\ddagger}$, retrained & \CrossDmrNtuOTAR & \CrossDmrNtuOTRI & -- & -- \\
PMR$^{\dagger\ddagger}$, retrained & \CrossPmrNtuOTAR & \CrossPmrNtuOTRI & -- & -- \\
DisentangledTMR$^\dagger$, retrained & \CrossTmrNtuOTAR & \CrossTmrNtuOTRI & \CrossTmrEtriAR & \CrossTmrEtriRI \\
\midrule
\mir{}, retrained (A2) & \CrossNtuOTRetAR & \CrossNtuOTRetRI & \CrossEtriRetAR & \CrossEtriRetRI \\
\mir{}, frozen SGN (A1) & \CrossNtuOTFrzAR & \CrossNtuOTFrzRI & \CrossEtriFrzAR & \CrossEtriFrzRI \\
\mir{}, frozen MixFormer (A1) & -- & \CrossNtuOTMixRI & \CrossEtriMixAR & \CrossEtriMixRI \\
\bottomrule
\end{tabular}
\end{table}

The NTU-120 frozen action checkpoints expose only 49 classes for 120 target actions; the ETRI MixFormer exposes 56 for 51. We omit these action scores rather than treat label-space mismatch as utility loss. On ETRI, the class-compatible frozen SGN RI is low (\CrossEtriFrzRI\%) but so is frozen AR (\CrossEtriFrzAR\%); we do not attribute this to any property of the ETRI population, since the experiment does not isolate a cause.

\section{Perturbation Profile and Qualitative Examples}
\label{app:qual}

\begin{figure}[!htbp]
\centering
\includegraphics[width=\columnwidth]{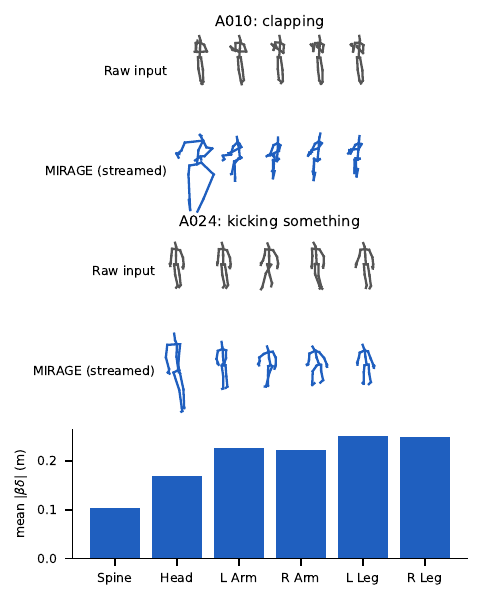}
\caption{Raw input and streamed \mir{} output for clapping and kicking: $x$--$y$ projections at five evenly spaced frames, with shared coordinate limits within each action. Bottom: mean perturbation by body part. The original retargeted sequences were unavailable for this visualization.}
\Description{Skeleton frames over time for raw input and MIRAGE for two actions, and a bar chart of mean perturbation magnitude by body part.}
\label{fig:qual}
\end{figure}

The mean per-joint magnitude of $\beta\boldsymbol\delta_t$ on a deterministic 5,000-clip subset of the generated NTU~60 train-and-test inventory is largest on the legs (\PertLegs\,m), smallest on the spine (\PertSpine\,m, a ratio of \PertLegSpineRatio), and intermediate on arms and head (\PertArmsHeadLo--\PertArmsHeadHi\,m). Clips are the first 5,000 sorted common filenames; raw frames undergo the same leading/trailing-zero trimming as generation. Each clip contributes equally. This is a descriptive subset, not a held-out population estimate. No body-part-specific perturbation targets are supplied. Figure~\ref{fig:qual} shows example frames.

\clearpage